%% file: main.tex
\documentclass[10pt,twocolumn,letterpaper]{article}

\usepackage[pagenumbers]{iccv} % To force page numbers, e.g. for an arXiv version

\input{preamble}

\definecolor{iccvblue}{rgb}{0.21,0.49,0.74}
\usepackage[pagebackref,breaklinks,colorlinks,allcolors=iccvblue]{hyperref}

\def\paperID{5636} % *** Enter the Paper ID here
\def\confName{ICCV}
\def\confYear{2025}

\title{4D Gaussian Splatting SLAM}

\author{ Yanyan Li, Youxu Fang,  Zunjie Zhu, Kunyi Li, Yong Ding, Federico Tombari \\
Technical University of Munich\\
Hangzhou Dianzi University\\
Zhejiang University\\
Google\\
{\tt\small Project Page: \url{https://github.com/yanyan-li/4DGS-SLAM}}
}

\begin{document}

\twocolumn[{%
    \renewcommand\twocolumn[1][]{#1}%
    \maketitle
    \begin{center}
        \centering
        \vspace{-15pt}
        
        % First row
        \begin{minipage}[b]{0.24\linewidth}
            \includegraphics[width=1\linewidth]{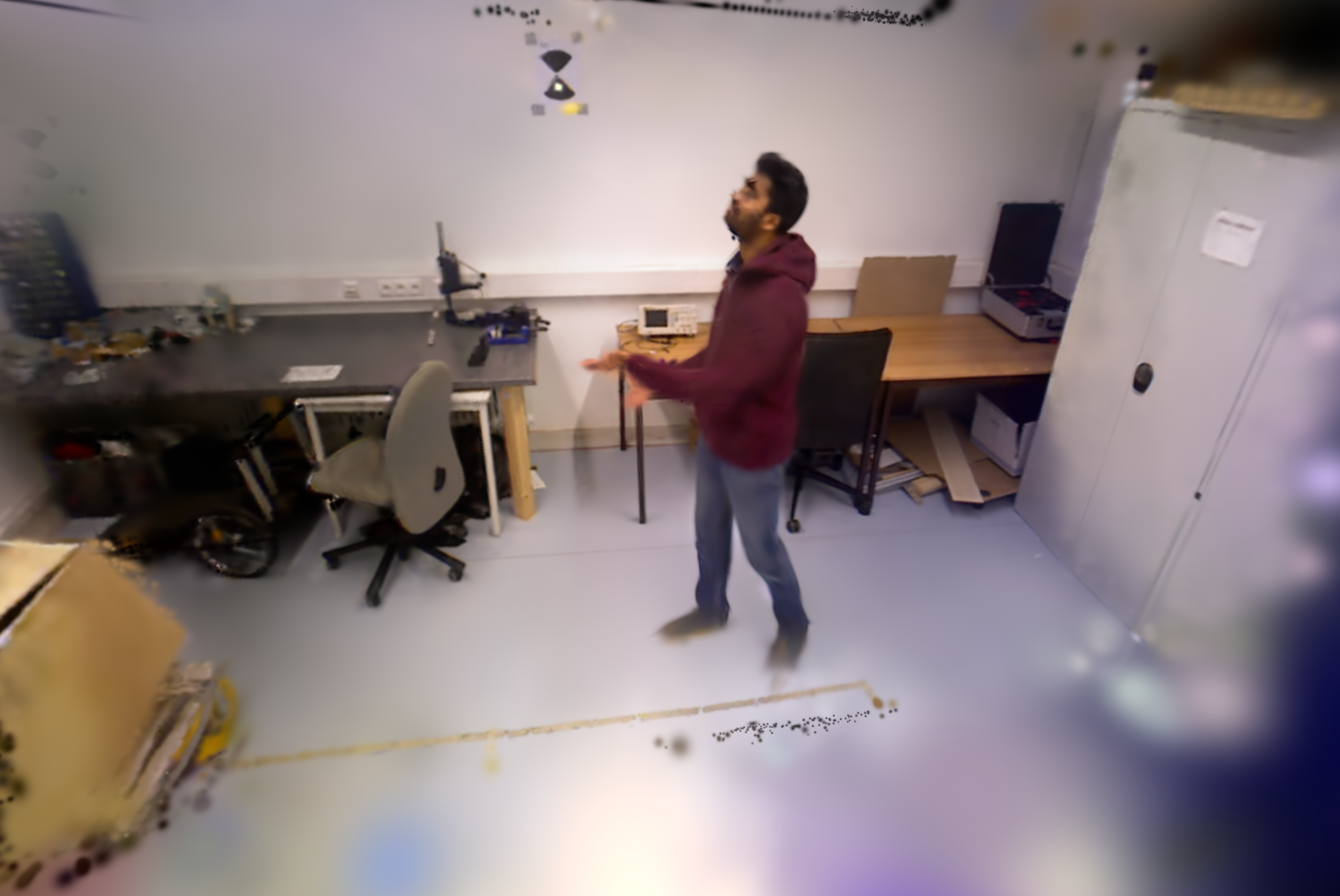}%
        \end{minipage}
        \hfill
        \begin{minipage}[b]{0.24\linewidth}
            \includegraphics[width=1\linewidth]{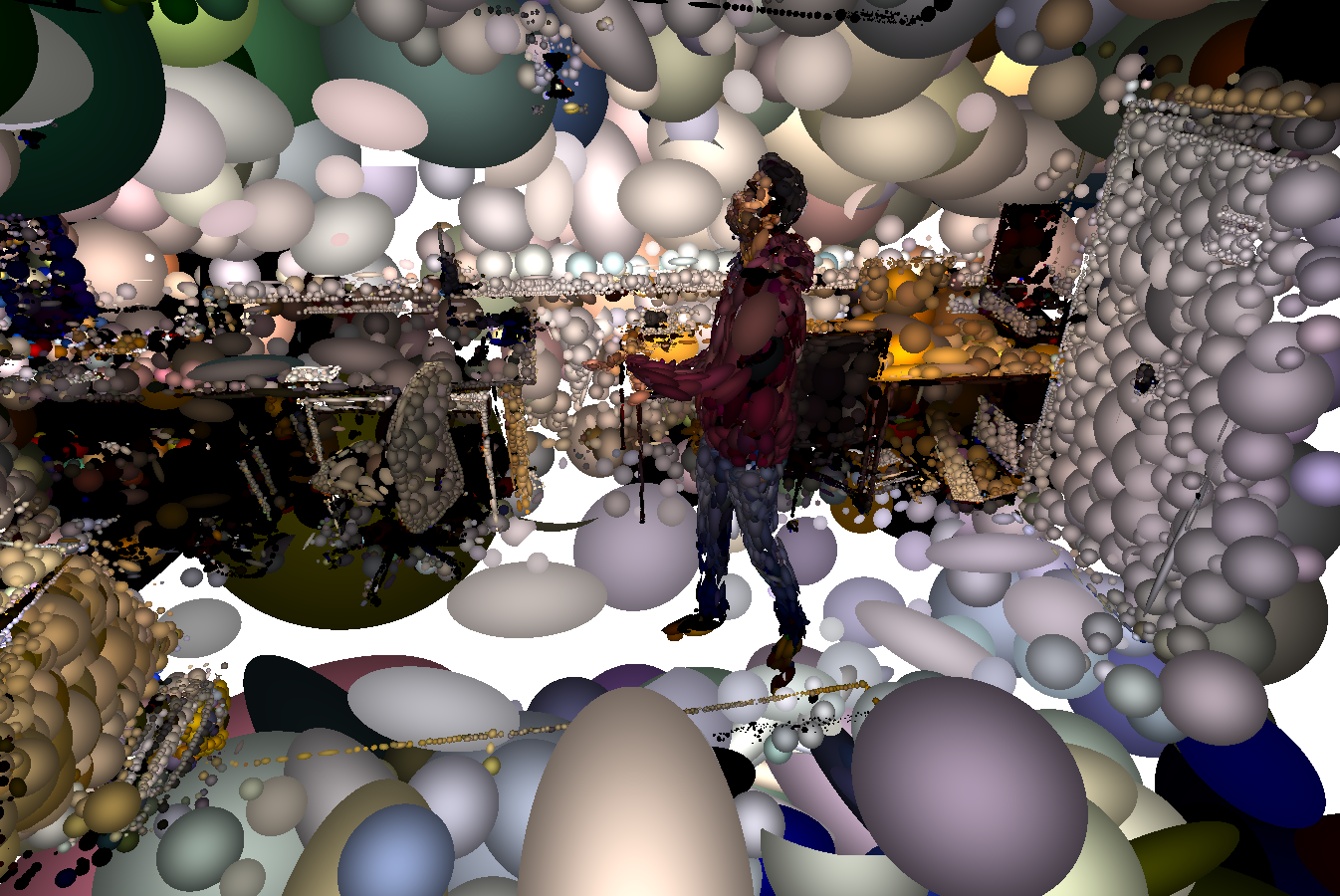}%
        \end{minipage}
        \hfill
        \begin{minipage}[b]{0.24\linewidth}
            \includegraphics[width=1\linewidth]{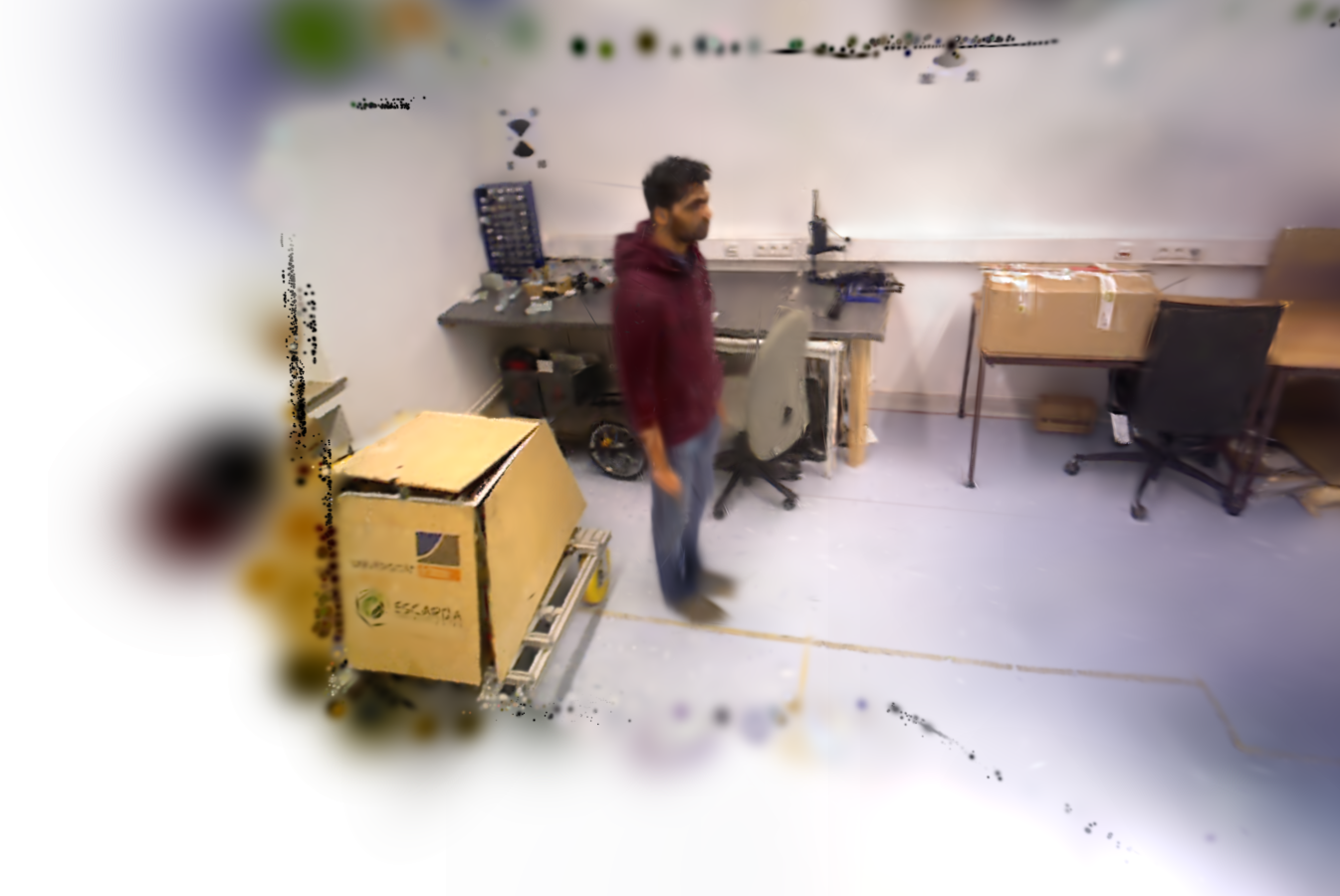}%
        \end{minipage}
        \hfill
        \begin{minipage}[b]{0.24\linewidth}
            \includegraphics[width=1\linewidth]{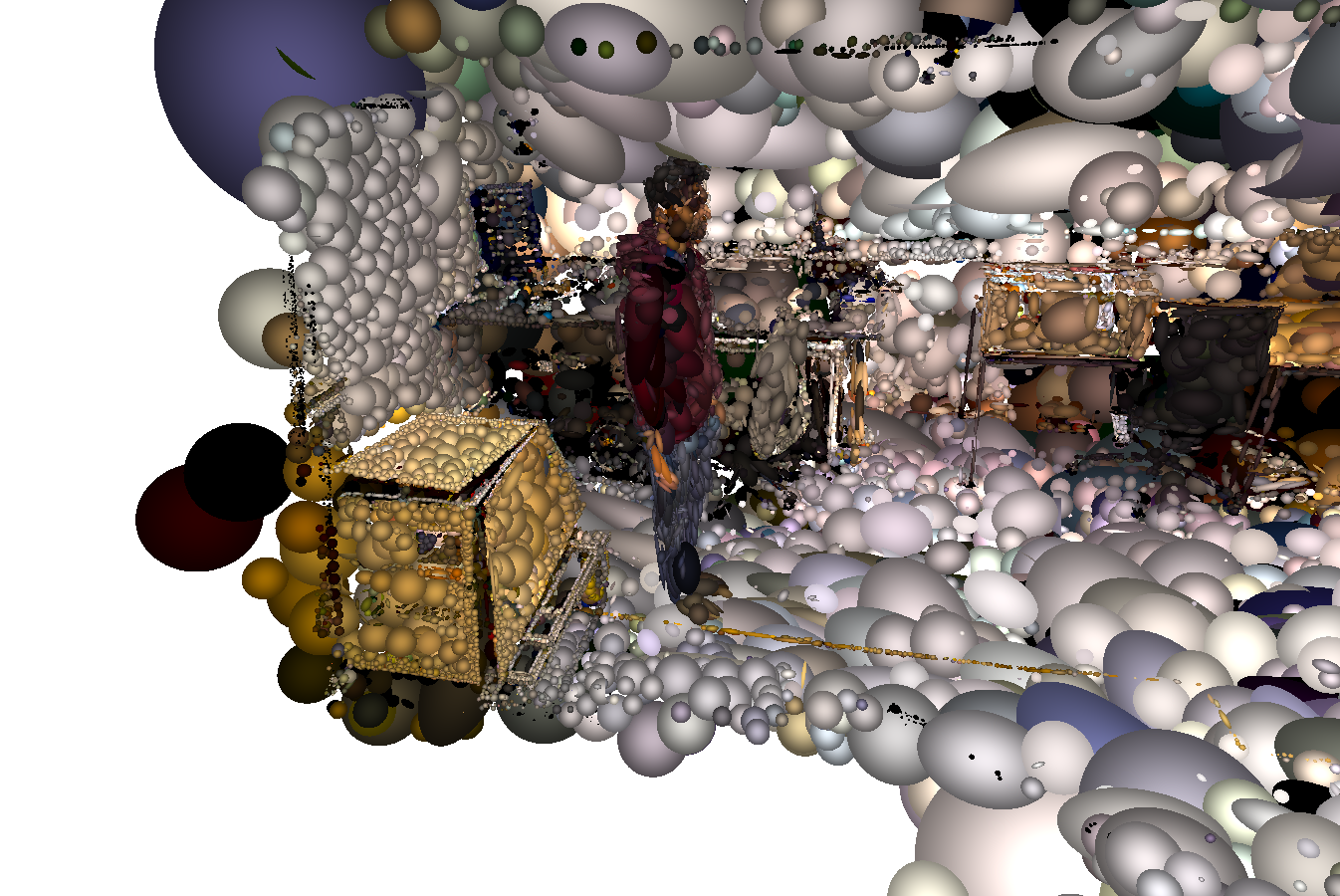}%
        \end{minipage}

        % Second row
        \vspace{1mm}

        \begin{minipage}[b]{0.24\linewidth}
            \includegraphics[width=1\linewidth]{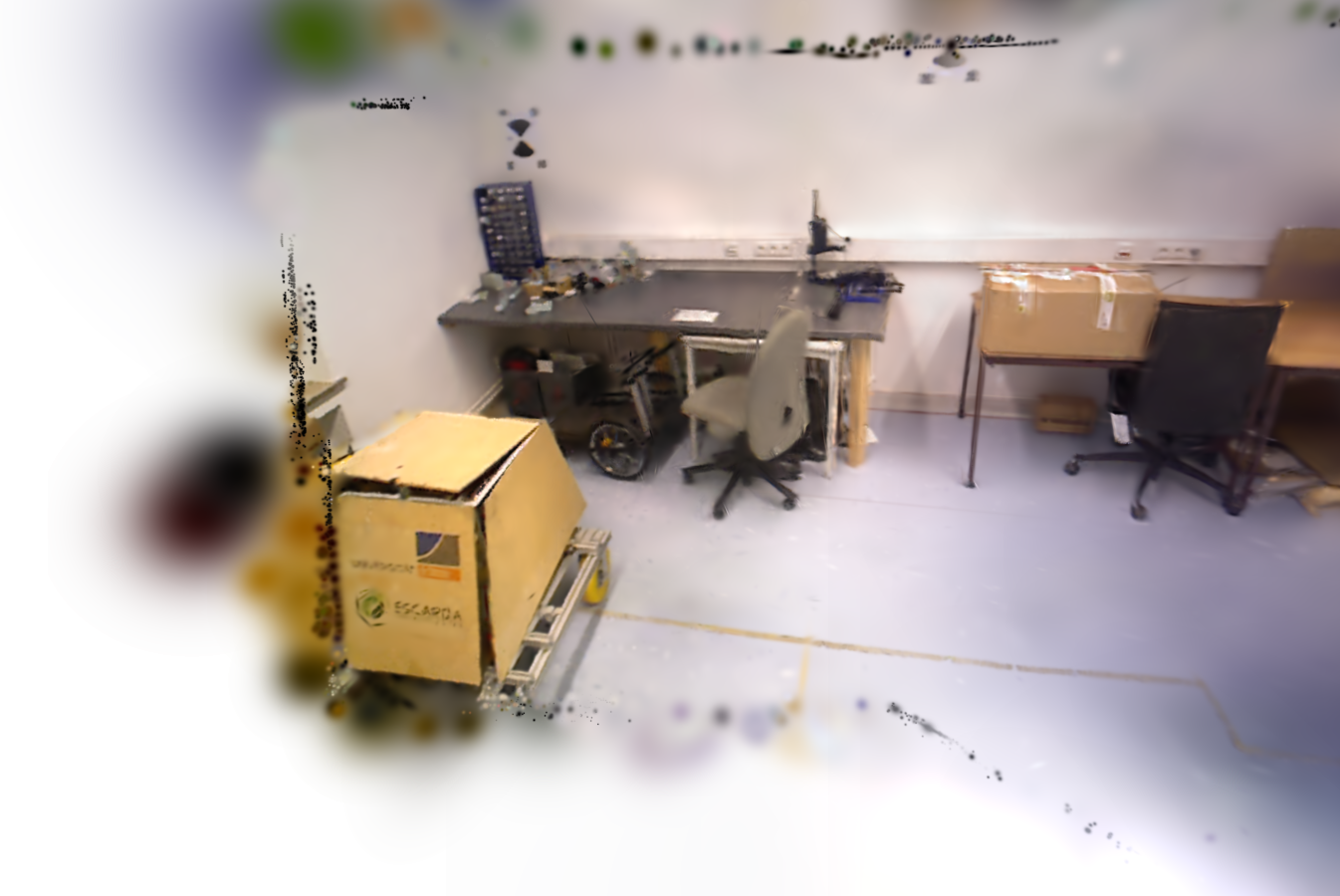}%
        \end{minipage}
        \hfill
        \begin{minipage}[b]{0.24\linewidth}
            \includegraphics[width=1\linewidth]{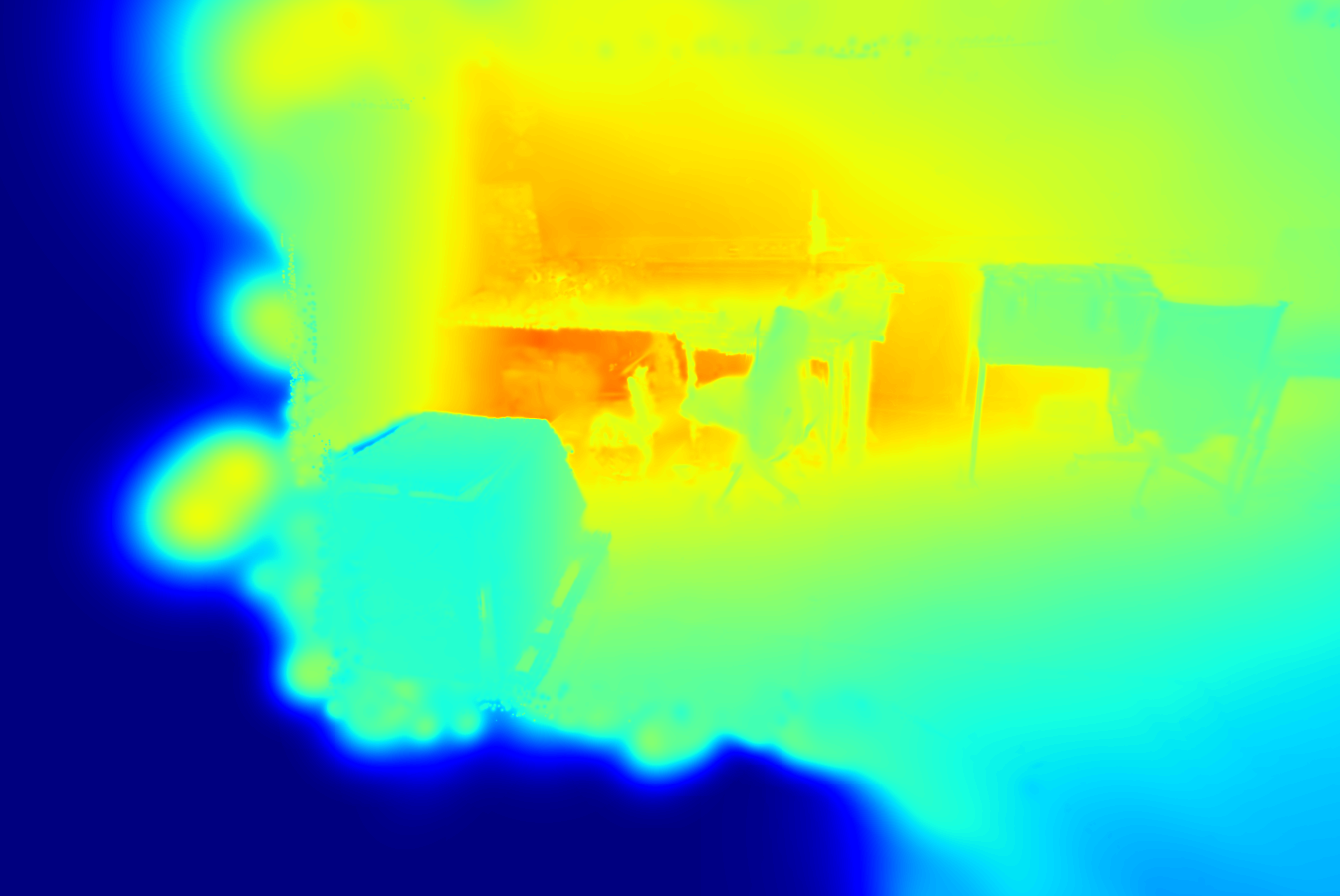}%
        \end{minipage}
        \hfill
        \begin{minipage}[b]{0.24\linewidth}
            \includegraphics[width=1\linewidth]{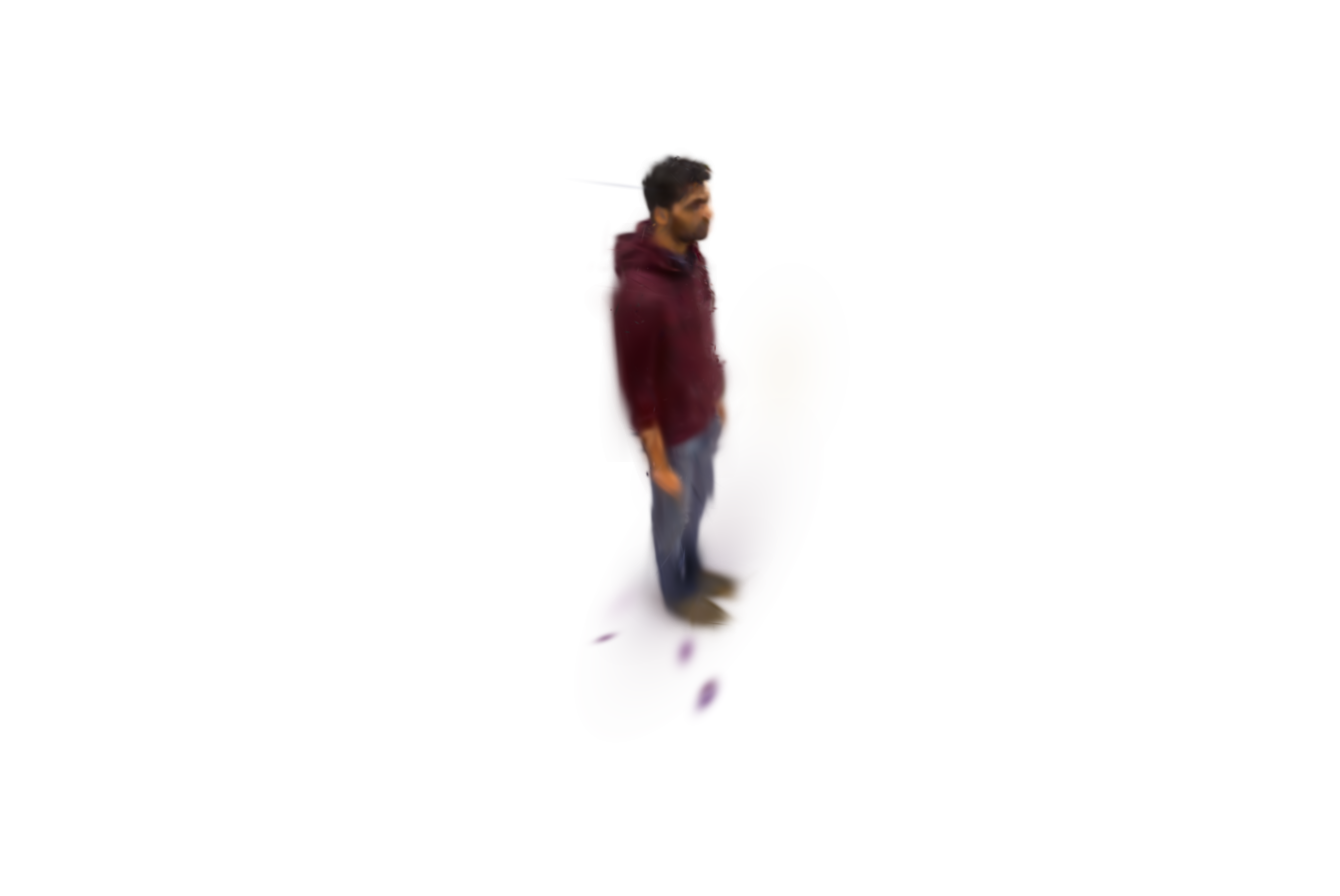}%
        \end{minipage}
        \hfill
        \begin{minipage}[b]{0.24\linewidth}
            \includegraphics[width=1\linewidth]{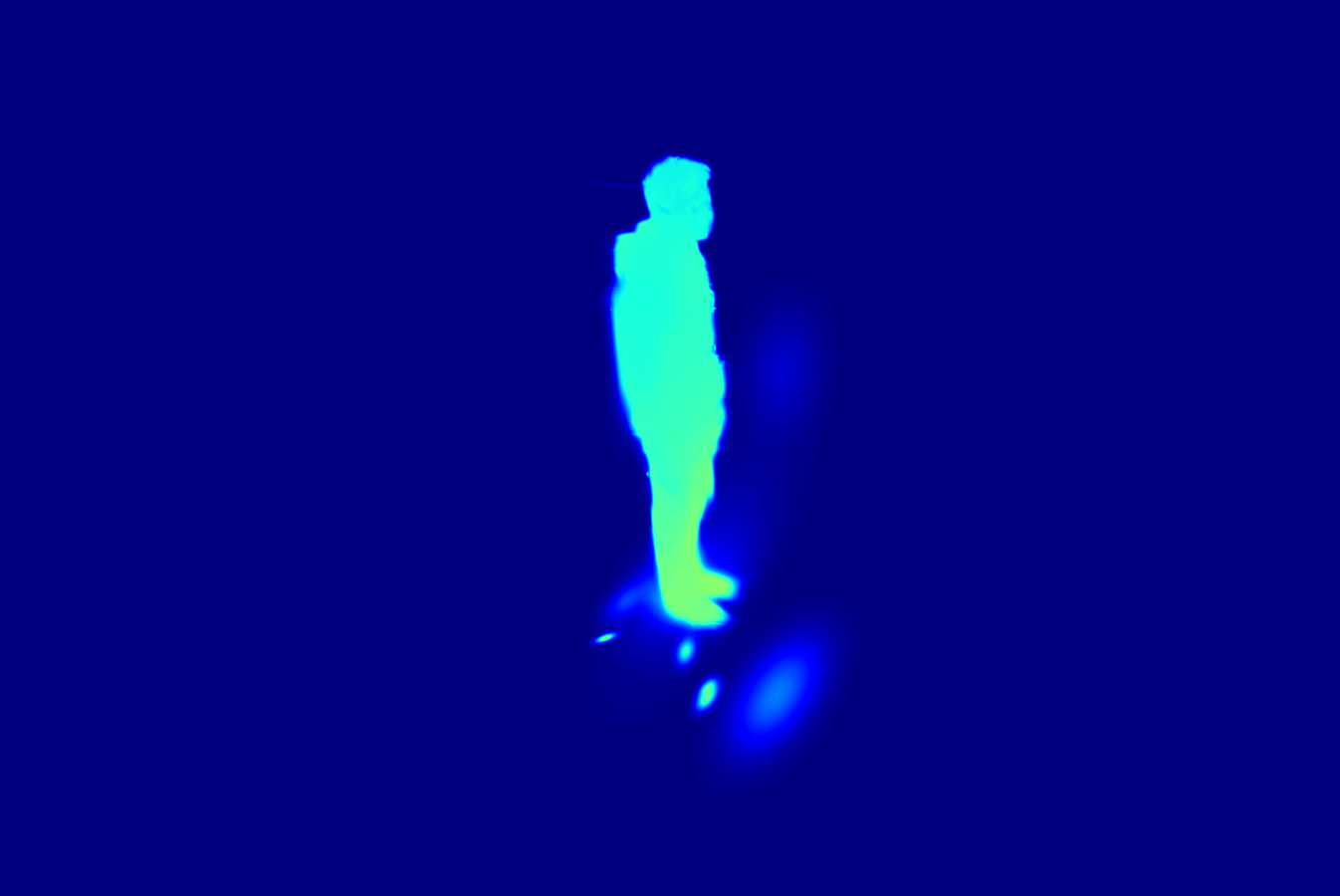}%
        \end{minipage}

        \captionof{figure}{\textbf{Example results from the proposed 4D-GS SLAM system. The top row showcases novel view synthesis and Gaussian visualizations in the BONN balloon (top left) and person\_tracking (top right) sequences. The appearance and geometry of static and dynamic scenes are shown in the bottom row, respectively.}}
        \label{BONN_vis_novel}
    \end{center}
}]

\input{sec/0_abstract}    
\input{sec/1_intro}

\input{sec/2_relatedwork}
\input{sec/3_method}

% \input{supp}

% \newpage
{
    \small
    \bibliographystyle{ieeenat_fullname}
    \bibliography{main}
}

% \input{sec/supp}

\end{document}

%% file: preamble.tex
\usepackage{bm}
\usepackage{graphicx}
\usepackage{multicol}
\usepackage{multirow}
\usepackage{subcaption}
\usepackage{bbding}

%% file: sec/0_abstract.tex
\begin{abstract}
% background
Simultaneously localizing camera poses and constructing Gaussian radiance fields in dynamic scenes establish a crucial bridge between 2D images and the 4D real world.
Instead of removing dynamic objects as distractors and reconstructing only static environments, this paper proposes an efficient architecture that incrementally tracks camera poses and establishes the 4D Gaussian radiance fields in unknown scenarios by using a sequence of RGB-D images.
% contribution 1
First, by generating motion masks, we obtain static and dynamic priors for each pixel.
% contribution 2
To eliminate the influence of static scenes and improve the efficiency on learning the motion of dynamic objects, we classify the Gaussian primitives into static and dynamic Gaussian sets, while the sparse control points along with an MLP is utilized to model the transformation fields of the dynamic Gaussians.
% contribution 3
To more accurately learn the motion of dynamic Gaussians, a novel 2D optical flow map reconstruction algorithm is designed to render optical flows of dynamic objects between neighbor images, which are further used to supervise the 4D Gaussian radiance fields along with traditional photometric and geometric constraints.
In experiments, qualitative and quantitative evaluation results show that the proposed method achieves robust tracking and high-quality view synthesis performance in real-world environments.
\end{abstract}

%% file: sec/1_intro.tex
\section{Introduction}
\label{sec:intro}

Tracking~\cite{mur2017orb,li2020structure}, mapping~\cite{izadi2011kinectfusion,fu2023colmap}, and rendering~\cite{kerbl3Dgaussians,rosinol2023nerf} in dynamic 3D scenes remain a fundamental challenge in computer vision, with important applications in robotics, augmented reality and autonomous systems.
% Tracking~\cite{mur2017orb,li2020structure}, mapping~\cite{izadi2011kinectfusion,fu2023colmap}, and rendering~\cite{kerbl3Dgaussians,rosinol2023nerf} in dynamic 3D scenes remain fundamental challenges in computer vision, with broad applications in robotics, augmented reality, and autonomous systems. 
While traditional methods~\cite{Matsuki:Murai:etal:CVPR2024,zhu2024robust,yugay2023gaussianslam} have demonstrated impressive localization and view synthesis capabilities in static environments, the presence of moving objects and diverse lighting conditions in real-world scenarios still significantly limit the performance of current solutions. 

3D Gaussian primitives~\cite{zwicker2002ewa,kerbl3Dgaussians} have recently emerged as a powerful representation for novel view synthesis and scene reconstruction, demonstrating efficient performance in training and rendering compared to Neural Radiance Field (NeRF) methods~\cite{barron2022mip,rosinol2023nerf}. However, pioneering Gaussian Splatting SLAM algorithms~\cite{Matsuki:Murai:etal:CVPR2024, ha2024rgbdgsicpslam} mostly assumed a static working space. Based on photometric and geometric constraints, these methods can incrementally localize camera poses and optimize Gaussian primitives in unknown scenes. To extend pose estimation capabilities from static scenes to dynamic ones, the most popular strategy~\cite{henein2020dynamic,yu2018ds} is to detect dynamic objects from 2D images and try to remove non-static pixels during the tracking process by leveraging semantic priors~\cite{hou20193d,kirillov2023segment}.

Following a similar dynamic detection strategy, dynamic Gaussian Splatting SLAM~\cite{xu2025dg,kong2024dgs} systems are proposed to extend the working fields to non-static environments. Based on the support of high-quality dynamic object detection methods~\cite{kirillov2023segment}, the localization accuracy is further improved also for dynamic Gaussian Splatting SLAM methods. However, after removing the detected dynamic pixel areas, current approaches fall back reconstructing static Gaussian radiance fields instead of building high-fidelity 4D reconstructions. 

% \begin{figure}
%     \centering
%     \includegraphics[width=\linewidth]{imgs/teaser.png}
%     \caption{Caption}
%     \label{fig:teaser}
% \end{figure}

\begin{figure*}[htpb]
    \centering
    \includegraphics[width=\linewidth]{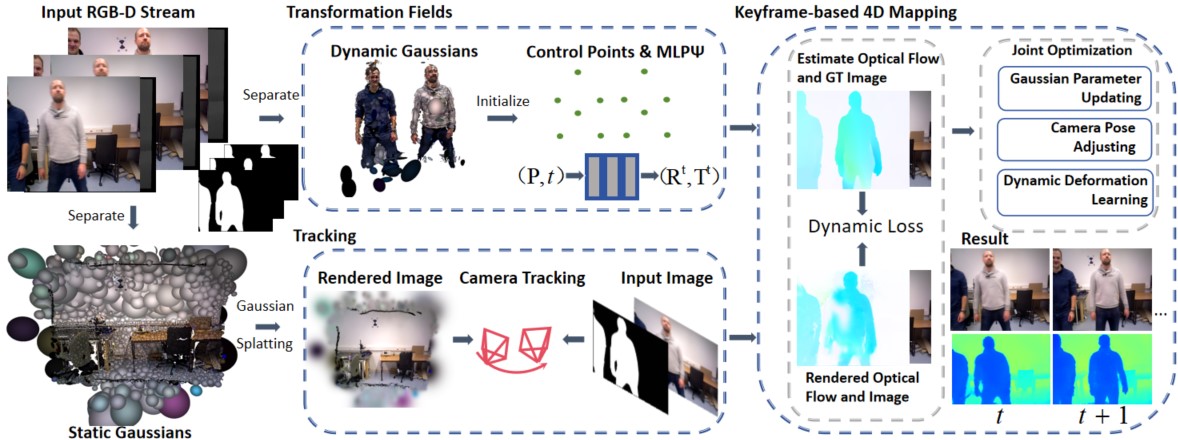}
    \caption{\textbf{Architecture of the proposed Gaussian Splatting SLAM.} The inputs to our system are temporally sequential RGB-D image sequences and motion masks.In the initial frame, dynamic and static Gaussians are independently initialized using a motion mask, and sparse control points are established according to the spatial distribution of dynamic Gaussians. The static structure is subsequently employed for camera pose estimation through photometric and geometric constraints. Following keyframe insertion, we co-optimize Gaussian attributes and camera poses while simultaneously estimating temporal motion patterns of dynamic Gaussians.}
    \label{fig:architecture}
\end{figure*}

To bridge this gap, we introduce a method that simultaneously localizes camera poses and reconstructs 4D Gaussian radiance fields from a sequence of RGB-D images in dynamic scenes. Instead of treating dynamic objects as noise~\cite{matsuki2024gaussian} or distractors~\cite{sabour2023robustnerf}, the proposed approach explicitly models temporal variations of the Gaussian radiance fields, enabling accurate scene representation while maintaining geometric consistency.
Our framework incrementally estimates camera poses and updates Gaussian representations in an online manner, ensuring robustness to unknown and highly dynamic environments. By leveraging depth information from RGB-D inputs, we improve geometric accuracy while maintaining efficient computation. Unlike prior work that relies on post-processing or explicit motion segmentation, our method naturally integrates motion cues into the scene representation, allowing for seamless reconstruction without discarding dynamic content. The contributions of our method can be summarized as follows:
\begin{itemize}
    \item A novel 4D Gaussian Splatting pipeline is proposed to localize camera poses and represent dynamic scenes in Gaussian radiance fields.
    % RGB-D SLAM method that combines dynamic Gaussian splatting and 3D Gaussian splatting can achieve reconstruction of both static scenes and dynamic objects.
    \item We divide the primitives into static and dynamic Gaussians and introduce sparse control points together with an MLP for modeling the motion of the dynamic Gaussians.
    \item A novel 2D optical flow rendering algorithm is proposed to improve the performance of 4D Gaussian fields.
    We estimate the 2D optical flow maps separately from dynamic GS and a pre-trained model, then leveraged as constraint to learn the motion of the dynamic Gaussians.
\end{itemize}

%% file: sec/2_relatedwork.tex
\section{Related Work}
\label{sec:related_work}
\paragraph{Camera Pose Estimation.}
Camera pose estimation is a fundamental task in communities of computer vision and robotics. Given monocular~\cite{mur2015orb,engel2014lsd}, stereo~\cite{mur2017orb,engel2015large}, RGB-D~\cite{li2021rgb,schops2019bad}, or visual-inertial~\cite{qin2018vins,campos2021orb}, popular algorithms in the domain of multiple view geometry are proposed to estimate translation and orientation matrices via 2D-2D~\cite{haralick1989pose,luong1997self}, 2D-3D~\cite{zheng2013revisiting,hesch2011direct}, and 3D-3D~\cite{segal2009generalized,rusinkiewicz2001efficient} strategies. Extended from these fundamental theories, robust and versatile systems~\cite{mur2015orb,dai2017bundlefusion,li2020structure,izadi2011kinectfusion} are implemented to obtain track cameras and reconstruct unknown environments. There are different focuses between these systems, where the first group~\cite{mur2015orb} of systems pursue accurate localization results while another type~\cite{dai2017bundlefusion} of pipelines achieve dense and high-quality 3D reconstructions. 
With the development of deep neural networks, deep point~\cite{detone2018superpoint} and line~\cite{zhang2019ppgnet} are used in feature matching. RAFT~\cite{teed2020raft} predicts optical flow maps between relative images. 

\paragraph{3D Gaussian Splatting and Non-static GS SLAM.}
3D Gaussian Splatting (3DGS)~\cite{kerbl3Dgaussians,li2024geogaussian} is an explicit parametrization for representing 3D unknown scenes, which shows more efficient performance than implicit methods, like NeRF~\cite{mildenhall2021nerf} in novel view rendering tasks. 
For traditional 3DGS methods\cite{Matsuki:Murai:etal:CVPR2024, keetha2024splatam, yugay2023gaussianslam}, the application fields mainly focus on static scenes. These approaches have demonstrated strong performance in environments where the scene remains largely unchanged over time, enabling accurate tracking and reconstruction of 3D structures. However, in dynamic scenes, these methods tend to incur significant errors during tracking or reconstruction. For non-static scenes, methods~\cite{xu2025dg,kong2024dgs,hou2025mvgsr} explore strategies to deal with dynamic objects as distractors and establish Gaussian fields for static components after removing dynamic objects. Compared to these non-static Gaussian Splatting methods that assume camera poses are given, non-static GS SLAM methods~\cite{xu2025dg,kong2024dgs} are incrementally fed by Monocular or RGB-D images to estimate camera poses and reconstruct Gaussian primitives. To achieve the goal, dynamic object instances are masked from 2D images based on semantic detection methods.  Furthermore, these removed regions are recovered by multiple views during the optimization process.  

\paragraph{Dynamic Gaussian Splatting.} Dynamic 3D Gaussian technology enhances the fast rendering capabilities of 3DGS~\cite{kerbl3Dgaussians}, adapting it for dynamic scene reconstruction. In this context, 4D Gaussian splatting~\cite{Wu_2024_CVPR} (4DGS) presents an innovative approach by combining 3D Gaussians with 4D neural voxels. It introduces a decomposition neural voxel encoding method, inspired by HexPlane~\cite{Cao2023HexPlane}, to efficiently generate Gaussian features from these 4D neural voxels. To handle temporal variations, a lightweight MLP is applied to predict Gaussian deformations over time. 
Building on this, the D3DGS framework~\cite{bae2024ed3dgs} offers a deformable 3DGS model for dynamic scene representation, where time is conditioned on the 3DGS. This framework transforms the learning process into a canonical space, allowing for the joint training of a purely implicit deformable field with the learnable 3DGS. The result is a time-independent 3DGS that separates motion from geometry.
Additionally, 3D Gaussians for Efficient Streaming~\cite{sun20243dgstream} significantly optimizes the streaming of photo-realistic Free-Viewpoint Videos (FVVs) for dynamic scenes. It achieves this by using a compact Neural Transformation Cache (NTC) to simulate the translation and rotation (transformation fields~\cite{hu2024learnable}) of 3D Gaussians. This method reduces the training time and storage space needed for each FVV frame while introducing an adaptive strategy to accommodate new objects in dynamic scenes.

%% file: sec/3_method.tex
\section{Methodology}

\subsection{Initialization}
%：
Similar to GS-based SLAM systems~\cite{Matsuki:Murai:etal:CVPR2024, keetha2024splatam, yugay2023gaussianslam}, the traditional components of 3D Gaussian ellipsoids, including mean $\bm{\mu}$, covariance $\bm{\Sigma}$, opacity $\alpha$, and color $\mathbf{c}$ parameters, are utilized in our representation. But the difference is that we further define new attribute $dy$ to each Gaussian, which is used to represent whether the Gaussian is dynamic Gaussian or not. Therefore, the final representation is $\mathcal{G}=[ \bm{\Sigma} \; \bm{\mu} \; \alpha \; \mathbf{c} \; dy ]$.  

Following 3D Gaussian Splatting~\cite{kerbl3Dgaussians}, each 3D Gaussian is rasterized into 2D splats, allowing for gradient flow in scene reconstruction and pose estimation. As a result, the rendered color of a pixel, denoted as $C(p)$, can be described by the following equation:
\begin{equation}
    C(p)=\sum_{i=1}^{n} c_i\alpha_i \prod_{j}^{i-1}(1-\alpha_j)
\end{equation}
here, $c$ and $\alpha$ are the color and opacity properties of the Gaussian, respectively.

Additionally, per-pixel depth D(p) and opacity O(p) are rasterized by using alpha-blending:
\begin{equation}
    D(p)=\sum_{i=1}^{n} d_i\alpha_i \prod_{j}^{i-1}(1-\alpha_j)
\end{equation}

\begin{equation}
    O(p)=\sum_{i=1}^{n} \alpha_i \prod_{j}^{i-1}(1-\alpha_j)
\end{equation}
where $d_i$ is the distance to the mean $\bm{\mu}$ of the $i^{th}$ Gaussian along the camera ray.
% \textcolor{red}{Motion mask? }
% Here the semantic model used in this paper is YoLov9~\cite{wang2024yolov9} to instance mask for dynamic objects.

% where each Gaussian  
% the proposed method utilizes  to represent scenarios
Instead of assuming that environments are static~\cite{Matsuki:Murai:etal:CVPR2024, keetha2024splatam, yugay2023gaussianslam} or removing dynamic objects~\cite{xu2025dg,kong2024dgs} in Gaussian Splatting optimization, we explore strategies to establish the dynamic deformation network for dynamic Gaussians. 
To be specific, we use a pre-trained model YoLov9~\cite{wang2024yolov9} to obtain the motion mask. For sequences containing dynamic objects that the pre-trained model cannot correctly segment, we generate the motion mask by combining optical flow and the pre-trained model.
Based on the detected dynamics, the Gaussians associated with pixels lying on the motion masks are defined as dynamic Gaussians ($\mathcal{G}_{dy}$), while others are initialized as static Gaussians ($\mathcal{G}_{st}$), during the initialization stage.

Inspired by SC-GS~\cite{huang2023sc}, we also make use of sparse control points to learn the 6 DoF transformation. However, the difference is that instead of obtaining sparse control points through long-term pre-training, we initialize these points using the motion regions from the input image of the initial frame.  

For each control point, we learn a time-varying 6-DoF transformation via an MLP $\Psi$ block. Therefore, the process of querying the transformation fields of each control point $P_k$ at each time step $t$, which can be denoted as:
\begin{equation}
    \Psi(P_k, t) \to [\mathbf{R}^t, \mathbf{T}^t].
\end{equation}

What is more, we derive the dense transformation field of dynamic Gaussians using local interpolation of the transformations of their neighboring control points, employing Linear Blend Skinning (LBS)~\cite{10.1145/1276377.1276478}. Specifically, for each dynamic Gaussian $\mathcal{G}_{dy}$, we use K-Nearest Neighbors (KNN) search to find its $K$ nearest control points ${p_k | k \in N_j}$ in the canonical space. 
Then, the interpolation weights for the control points $p_k$ can be computed using a Gaussian Radial Basis Function (RBF). By using the interpolation weights of the neighboring control points and the 6-DoF transformations, we can compute the scale $\mathbf{S}$, rotation $\mathbf{R}$, and positional $\bm{\mu}$ changes of each dynamic Gaussian $\mathcal{G}_{dy}$.

\subsection{Tracking}
To avoid interference from the motion of dynamic objects in the input and rendered images on camera tracking, we exclude dynamic Gaussians from the Gaussian splatter rendering during the tracking process. Instead, we optimize the camera pose and exposure parameters using the rendered color and depth maps, which are generated only by static Gaussians. The optimization is performed using $L_1$ loss between the rendered appearance and depth maps and their observations, where the motion mask $\mathcal{M}$ is used here to remove dynamic objects from the input images to achieve robust camera pose localization performance.
% The specific loss calculation is the same as that in MonoGS~\cite{Matsuki:Murai:etal:CVPR2024}, except that, outside the motion mask region, the depth $L_1$ loss mask is defined as opacity image $O(p) > 0.95$ and depth $ d> 0$, and the color $L_1$ loss mask is defined as areas where the gradient of the color image exceeds a certain threshold. The final loss is:
\begin{equation}
    L_t = \sum_p \mathcal{M}(\lambda O(p)L_1(C(p))+ (1 -\lambda)L_1(D(p)))
\label{eq:pose_localize}
\end{equation}
here, an L1 loss is to supervise both the depth and color renders, and $\lambda$ are fixed weights during the optimization process.
Note that, for $L_1(D(p))$, we only apply the loss over pixels that $O(p)>0.95$ and the ground-truth depth $d(p)>0$. For $L_1(C(p))$, we only apply the loss over pixels that gradient of the ground-truth color image exceeds a certain threshold $\sigma$.

% bonn ATE comparison 
\begin{table*}[ht]
    \centering
    \renewcommand{\arraystretch}{1.2} %rows, default value is 1.0
    %\captionsetup[table*]{singlelinecheck=off}
	\setlength{\tabcolsep}{5pt}
    \resizebox{\textwidth}{!}{
     \begin{tabular}{c|cccccccccc}
        \toprule
        Method           & \multicolumn{1}{l}{ballon} & \multicolumn{1}{l}{ballon2} & \multicolumn{1}{l}{ps\_track} & \multicolumn{1}{l}{ps\_track2} & \multicolumn{1}{l}{sync} & \multicolumn{1}{l}{sync2} & \multicolumn{1}{l}{p\_no\_box}& \multicolumn{1}{l}{p\_no\_box2}& \multicolumn{1}{l}{p\_no\_box3}& \multicolumn{1}{l}{Avg.} \\
        \hline
        \multirow{1}{*}{RoDyn-SLAM\cite{jiang2024rodynslam}}  &7.9&  11.5&  14.5  & 13.8  &  \textbf{1.3}     & 1.4  & 4.9 & 6.2 & 10.2 & 7.9 \\
        \hline
        \multirow{1}{*}{MonoGS\cite{Matsuki:Murai:etal:CVPR2024}}  & 29.6 & 22.1   & 54.5  & 36.9   & 68.5  & 0.56  & 71.5 & 10.7 & 3.6 & 33.1 \\
        \hline
        \multirow{1}{*}{Gaussian-SLAM\cite{yugay2023gaussianslam}}  & 66.9 & 32.8   & 107.2& 114.4   & 111.8  & 164.8  & 69.9 & 53.8 & 37.9 & 84.3\\
        \hline
        \multirow{1}{*}{SplaTAM\cite{keetha2024splatam}}  & 32.9 & 30.4   & 77.8  & 116.7   & 59.5  & 66.7  & 91.9 & 18.5 & 17.1 & 56.8 \\
        \hline
        \multirow{1}{*}{Ours}  &\textbf{2.4}    & \textbf{3.7}   & \textbf{8.9}     & \textbf{9.4}   & 2.8  & \textbf{0.56}  & \textbf{1.8} & \textbf{1.5} & \textbf{2.2} & \textbf{3.6} \\
    \bottomrule
    \end{tabular}}
\caption{\textbf{Trajectory errors in ATE [cm]$\downarrow$ in the BONN sequences.} Results with the best accuracy are highlighted in \textbf{bold} font.}
\label{BONN ATE}
\end{table*}

% tum拟数据集PSNR对比 
\begin{table*}[ht]
    \centering
    \captionsetup[table*]{singlelinecheck=off}
	\renewcommand{\arraystretch}{1.2} %rows, default value is 1.0
	\setlength{\tabcolsep}{9pt}
        \resizebox{\textwidth}{!}{
        \begin{tabular}{c|c|ccccccc}
        \toprule
        Method                  & Metric            & \multicolumn{1}{l}{fr3/sit\_st} & \multicolumn{1}{l}{fr3/sit\_xyz} & \multicolumn{1}{l}{fr3/sit\_rpy} & \multicolumn{1}{l}{fr3/walk\_st} & \multicolumn{1}{l}{fr3/walk\_xyz} & \multicolumn{1}{l}{fr3/walk\_rpy} & \multicolumn{1}{l}{Avg.} \\
        \hline
        \multirow{3}{*}{MonoGS\cite{Matsuki:Murai:etal:CVPR2024}} & 
        PSNR{[}dB{]} $\uparrow$  & 19.95  & 23.92  & 16.99  & 16.47   & 14.02   & 15.12  & 17.74  \\
        & SSIM$\uparrow$   & 0.739   & 0.803   & 0.572  & 0.604   & 0.436   & 0.497  & 0.608       \\
        & LPIPS$\downarrow$  & 0.213  & 0.182  & 0.405    & 0.355   & 0.581    & 0.56   & 0.382   \\
        \hline
        \multirow{3}{*}{Gaussian-SLAM\cite{yugay2023gaussianslam}} & 
        PSNR{[}dB{]} $\uparrow$  & 18.57  & 19.22  & 16.75  & 14.91   & 14.67   & 14.5  & 16.43  \\
        & SSIM$\uparrow$   & 0.848   & 0.796   & 0.652 & 0.607   & 0.483   & 0.467  & 0.642       \\
        & LPIPS$\downarrow$  & 0.291  & 0.326  & 0.521   & 0.489   & 0.626    & 0.630   & 0.480   \\
        \hline
        \multirow{3}{*}{SplaTAM\cite{keetha2024splatam}} & 
        PSNR{[}dB{]} $\uparrow$  & 24.12  & 22.07  & 19.97  & 16.70   & 17.03   & 16.54  & 19.40  \\
        & SSIM$\uparrow$   & \textbf{0.915}   & \textbf{0.879}   & \textbf{0.799} & 0.688   & 0.650   & 0.635  & 0.757       \\
        & LPIPS$\downarrow$  & \textbf{0.101}  & \textbf{0.163}  & \textbf{0.205}    & 0.287   & 0.339    & 0.353   & 0.241   \\
        \hline
        \multirow{3}{*}{SC-GS\cite{huang2023sc}} & 
        PSNR{[}dB{]} $\uparrow$  & 27.01  & 21.45  & 18.93  & 20.99   & \textbf{19.89}   & 16.44  & 20.78  \\
        & SSIM$\uparrow$   & 0.900   & 0.686   & 0.529 & 0.762   & 0.590   & 0.475  & 0.657       \\
        & LPIPS$\downarrow$  & 0.182  & 0.369  & 0.512    & 0.291   & 0.470    & 0.554   & 0.396   \\
        \hline
        \multirow{3}{*}{Ours} & 
        PSNR{[}dB{]} $\uparrow$  & \textbf{27.68}  & \textbf{24.37}  & \textbf{20.71}  & \textbf{22.99}   & 19.83   & \textbf{19.22}  & \textbf{22.46}  \\
        & SSIM$\uparrow$   & 0.892   & 0.822   & 0.746 & \textbf{0.820}   & \textbf{0.730}   & \textbf{0.708}  & \textbf{0.786}       \\
        & LPIPS$\downarrow$  & 0.116  & 0.179  & 0.265    & \textbf{0.195}   & \textbf{0.281}    & \textbf{0.337}   & \textbf{0.228}   \\
        \bottomrule
        \end{tabular}}
\caption{ \textbf{Quantitative results in the TUM RGB-D sequences.} Results with the best accuracy are highlighted in \textbf{bold} font. }
\label{TUM render}
\end{table*}
% bonn拟数据集PSNR对比 
\begin{table*}[ht]
    \centering
    \captionsetup[table*]{singlelinecheck=off}
	\renewcommand{\arraystretch}{1.2} %rows, default value is 1.0
	\setlength{\tabcolsep}{4pt}
        \resizebox{\textwidth}{!}{
        \begin{tabular}{c|c|cccccccccc}
        \toprule
        Method                  & Metric            & \multicolumn{1}{l}{ballon} & \multicolumn{1}{l}{ballon2} & \multicolumn{1}{l}{ps\_track} & \multicolumn{1}{l}{ps\_track2} & \multicolumn{1}{l}{sync} & \multicolumn{1}{l}{sync2} & \multicolumn{1}{l}{p\_no\_box}& \multicolumn{1}{l}{p\_no\_box2}& \multicolumn{1}{l}{p\_no\_box3}& \multicolumn{1}{l}{Avg.} \\
        \hline
        \multirow{3}{*}{MonoGS\cite{Matsuki:Murai:etal:CVPR2024}} & 
        PSNR{[}dB{]} $\uparrow$  & 21.35  & 20.22  & 20.53  & 20.09   & 22.03   & 20.55  & 20.764 & 19.38 & 24.81 & 21.06  \\
        & SSIM$\uparrow$   & 0.803   & 0.758   & 0.779  & 0.718   & 0.766   & 0.841  & 0.748 & 0.753 & 0.857 & 780      \\
        & LPIPS$\downarrow$  & 0.316  & 0.354  & 0.408    & 0.426   & 0.328    & 0.5210   & 0.428 & 0.372 & 0.243 & 0.342   \\
        \hline
        \multirow{3}{*}{Gaussian-SLAM\cite{yugay2023gaussianslam}} & 
        PSNR{[}dB{]} $\uparrow$  & 20.45  & 18.55  & 19.60  & 19.09   & 21.04   & 21.35  & 19.99 & 20.35 & 21.22 & 20.18\\
        & SSIM$\uparrow$   & 0.792   & 0.718   & 0.744 & 0.719   & 0.784   & 0.837  & 0.750 & 0.768 & 0.814  & 0.769     \\
        & LPIPS$\downarrow$  & 0.457  & 0.480  & 0.484   & 0.496   & 0.402    & 0.364   & 0.509 & 0.493 & 0.441 & 0.458   \\
        \hline
        \multirow{3}{*}{SplaTAM\cite{keetha2024splatam}} & 
        PSNR{[}dB{]} $\uparrow$  & 19.65  & 17.67  & 18.30  & 15.57   & 19.33   & 19.67  & 20.81 & 21.69 & 21.41 & 19.34 \\
        & SSIM$\uparrow$   & 0.781   & 0.702   & 0.670 & 0.606   & 0.776   & 0.730  & 0.824 & 0.852 & 0.873  & 0.757     \\
        & LPIPS$\downarrow$  & 0.211  & 0.280  & 0.283    & 0.331   & 0.227    & 0.258   & 0.191 & 0.165 & 0.152 & 0.233  \\
        \hline
        \multirow{3}{*}{SC-GS\cite{huang2023sc}} & 
        PSNR{[}dB{]} $\uparrow$  & 22.3  & 21.38 & - & - & \textbf{23.62}  & 22.74   & 20.60   & 21.55  &19.24 & 21.63 \\
        & SSIM$\uparrow$   & 0.737   & 0.708 & - & - & 0.788 & 0.801   & 0.688   & 0.722  &0.628  & 0.724       \\
        & LPIPS$\downarrow$  & 0.448  & 0.450 &- &- & 0.427    & 0.359   & 0.515    & 0.491   &0.539 &0.461   \\
        \hline
        \multirow{3}{*}{Ours} & 
        PSNR{[}dB{]} $\uparrow$  & \textbf{25.90}  & \textbf{22.71}  & \textbf{21.78}  & \textbf{20.65}   & 23.25   & \textbf{25.42}  & \textbf{23.14} & \textbf{24.28}& \textbf{25.88}  &\textbf{23.66}\\
        & SSIM$\uparrow$   & \textbf{0.874}   & \textbf{0.838}   & \textbf{0.832} & \textbf{0.820}   & \textbf{0.812}   & \textbf{0.892}  & \textbf{0.845} & \textbf{0.873} & \textbf{0.886} &\textbf{0.852}    \\
        & LPIPS$\downarrow$  & \textbf{0.234}  & \textbf{0.264}  & \textbf{0.289}    & \textbf{0.294}   & \textbf{0.250}    & \textbf{0.169}   & \textbf{0.239}  &\textbf{0.224} &\textbf{0.207} &\textbf{0.241} \\
        \bottomrule
        \end{tabular}}
\caption{ \textbf{Quantitative results in the BONN sequences.} Results with the best accuracy are highlighted in \textbf{bold} font. And "-" means that reconstruction failure.} 
\label{BONN render}
\end{table*}

\begin{table*}[ht]
    \centering
    \renewcommand{\arraystretch}{1.2} %rows, default value is 1.0
    %\captionsetup[table*]{singlelinecheck=off}
	\setlength{\tabcolsep}{9pt}
    \resizebox{\textwidth}{!}{
    \begin{tabular}{c|ccccccc}
    \toprule
        Method  & \multicolumn{1}{l}{fr3/sit\_st} & \multicolumn{1}{l}{fr3/sit\_xyz} & \multicolumn{1}{l}{fr3/sit\_rpy} & \multicolumn{1}{l}{fr3/walk\_st} & \multicolumn{1}{l}{fr3/walk\_xyz} & \multicolumn{1}{l}{fr3/walk\_rpy} & \multicolumn{1}{l}{Avg.} \\
        \hline
        \multirow{1}{*}{RoDyn-SLAM\cite{jiang2024rodynslam}}  & 1.5 &  5.6  & 5.7  & 1.7  & 8.3  & 8.1 & 5.1 \\
        \hline
        \multirow{1}{*}{MonoGS\cite{Matsuki:Murai:etal:CVPR2024}}  & \textbf{0.48} & 1.7   & 6.1  & 21.9   & 30.7  & 34.2  & 15.8 \\
        \hline
        \multirow{1}{*}{Gaussian-SLAM\cite{yugay2023gaussianslam}}  & 0.72 & \textbf{1.4}   & 21.02& 91.50   & 168.1  & 152.0  & 72.4 \\
        \hline
        \multirow{1}{*}{SplaTAM\cite{keetha2024splatam}}  & 0.52 & 1.5   & 11.8  & 83.2   & 134.2  & 142.3  & 62.2 \\
        \hline
        \multirow{1}{*}{Ours}  &0.58    & 2.9   & \textbf{2.6}     & \textbf{0.52}   & \textbf{2.1}  & \textbf{2.6}  & \textbf{1.8} \\
    \bottomrule
    \end{tabular}}
\caption{\textbf{Trajectory errors in ATE [cm]$\downarrow$ in the TUM RGB-D sequences.} Results with the best accuracy are highlighted in \textbf{bold} font.}
\label{TUM ATE}
\end{table*}

\paragraph{Keyframe Selection.}
% The visibility between the current frame (i) and the previous keyframe (j) is measured by the Intersection over Union (IoU) of their Gaussian distributions. If the visibility is below a certain threshold or the relative displacement ($t_{ij}$) is too large compared to the median depth, the current frame (i) will be registered as a keyframe.
Similar to MonoGS~\cite{Matsuki:Murai:etal:CVPR2024}, we also maintain a small number of keyframes in the sliding window $W$, using visibility checks and translation thresholds to select keyframes, removing them if their overlap with the latest keyframe drops below a threshold. However, a new strategy, different from MonoGS~\cite{Matsuki:Murai:etal:CVPR2024}, is proposed by considering dynamic situations. Specifically, even if the camera movement is small, a new keyframe can also be selected and inserted when we detect the motion mask has a big difference or at least every $5$ frame. After adding a keyframe, we initialize new static Gaussians with the static part of the input image pixels from the current frame, followed by the mapping step. However, new dynamic Gaussians will not be added.

\subsection{4D Mapping}

Once new static and dynamic scenarios are inserted into the system after the tracking process, we propose a 4D mapping module to optimize the dynamic Gaussian radiance fields. 
%
% Gaussian paramters, control points, mlp, camera poses in the window. 
% two stages:
% optimize camera poses and mlp network. 
% dynamic mask, we increase the l1- weight.
% optical flow:
% deep model\cite{teed2020raftrecurrentallpairsfield} is used to predict the forward and backward optical map. feed last keyframe and current keyframe. 
% the process of rendering optical map: 
% dynamic gaussians states at different timestamp can be render two pixels, then we compute the distance map between.  

\paragraph{Optical Flow Map Rendering.}
As introduced in Equation~\ref{eq:pose_localize},
appearance (RGB) and geometry (depth) rendering constraints are utilized in the tracking process. However, in the 4D mapping section, these traditional single-view supervisions can provide reliable constraints for dynamic scenarios incrementally.  

To solve the problem, we are the first 4D Gaussian Splatting SLAM system that provides a novel strategy to render another type of map, Optical Flow Map, in the 4D mapping module. 
% steps
First of all, to create accurate optical flows between two images, the traditional methods~\cite{fleet2006optical} are using pixel-based tracking methods. Instead of from the perspective of 2D views and correspondence matching, we migrate the dynamic Gaussians $\mathcal{G}_{dy}$ between the currently selected keyframe and its last keyframe to obtain two corresponding sets of Gaussians, $G_t$ and $G_{t-1}$. These two sets of Gaussians are projected onto the camera plane of the current keyframe, resulting in two sets of 2D point coordinates $\mathbf{p}_t$ and $\mathbf{p}_{t-1}$. Let the difference between $p_t$ and $p_{t-1}$ be denoted as $d_x$. Similar to rendering color and depth maps, we can use $d_x$ to render the backward optical flow map $F(p)$ from time $t$ to $t-1$:
\begin{equation}
    F(p)=\sum_{i=1}^{n} d_x\alpha_i \prod_{j}^{i-1}(1-\alpha_j).
\end{equation}

Similarly, we can also render the forward optical flow map from frames $\textit{I}_{t-1}$ to $\textit{I}_t$. The optical flow loss is computed by comparing the forward and backward optical flow maps rendered from the dynamic Gaussians with the forward and backward optical flow maps estimated by RAFT~\cite{teed2020raftrecurrentallpairsfield} for the real input color images at times $t-1$ and $t$ in the motion mask area, using $L1$ loss, which can be denoted as:
\begin{equation}
    \begin{aligned}
    \mathcal{L}_{flow} =\ &  \sum_p \mathcal{M}(L_1(F(p)_{t \rightarrow t-1}, RAFT(p)_{t \rightarrow t-1})\\
                &+ L_1(F(p)_{t-1\rightarrow t}, RAFT(p)_{t-1\rightarrow t}))
    \end{aligned}
\end{equation}
here, $F(p)_{t \rightarrow t-1}$ and $F(p)_{t-1 \rightarrow t}$ are the optical flow maps of dynamic Gaussian rendering from time t to t-1 and from time t-1 to t, $RAFT(p)_{t \rightarrow t-1}$ and $RAFT(p)_{t-1 \rightarrow t}$ are the optical flow map estimated by RAFT~\cite{teed2020raftrecurrentallpairsfield} from time $t$ to $t-1$ and time $t-1$ to $t$.

 \paragraph{Joint Optimization}
In the mapping process, we use the first three keyframes in $W$ and randomly select five keyframes that overlap with the current frame to reconstruct the currently visible area. Additionally, to prevent forgetting the global map, two keyframes are randomly selected during each iteration. We optimize the Gaussian parameters and the camera poses of the three most recently added keyframes using the photometric $\mathcal{L}_{1}(C(p))$, geometric $\mathcal{L}_{1}(D(p))$. 

And we also introduce the regularization $\mathcal{L}_{iso}$ loss functions to penalize the  stretch of the ellipsoid $s_i$ by its difference to the mean $\tilde{s_i}$:
\begin{equation}
    E_{iso} = \Sigma_{i=1}^{|\mathcal{G}|}||s_i-\tilde{s_i}||_1
\end{equation}

Furthermore, we optimize the dynamic deformation network, which includes the MLP layers $\Psi$ and the parameters of control points. To achieve this, we also need to compute the ARAP loss~\cite{huang2023sc} and the optical flow loss for each map keyframe.
 
Finally, we optimize the relevant parameters by using a weighted sum of these losses, denoted as $L_{mapping}$.
\begin{equation}
    \begin{aligned}
    L_{mapping}=\ & \lambda L_1(C(p))+(1-\lambda)L_1(D(p)) \\
                &+\lambda_{flow} \mathcal{L}_{flow}+W_1arap\_{loss}\\
                & +W_2E_{iso}
    \end{aligned}
\end{equation}
here, $\lambda$, $\lambda_{flow}$, $W1$ and $W2$ are fixed weights during optimization.

Therefore, the two-stage mapping strategy is introduced to optimize the camera poses, exposure parameters, and dynamic deformation network. This strategy can be described in detail as follows:
 \begin{itemize}
     \item In the first stage, we use loss mapping $L_{mapping}$ to optimize only the camera poses and exposure parameters for the first three keyframes in $W$,  as well as the dynamic deformation network, without optimizing the Gaussian parameters. During this stage, the weight of the L1 loss for the color and depth maps in the motion mask region will be doubled.
     \item In the second stage, we use $L_{mapping}$ to optimize the camera poses and exposure parameters for the first three keyframes in $W$, dynamic deformation network, and Gaussian parameters.
 \end{itemize}

%TUM数据集渲染结果对比
\begin{figure*}[htbp]
	\centering
    \captionsetup[subfloat]{labelformat=empty}
	\subfloat{%
        \hspace{-5mm}%
        \rotatebox{90}{\scriptsize{~~~~~~~~\textbf{fr3\_sitting\_rpy}}}
		\begin{minipage}[b]{0.20\linewidth}
			\includegraphics[width=1\linewidth]{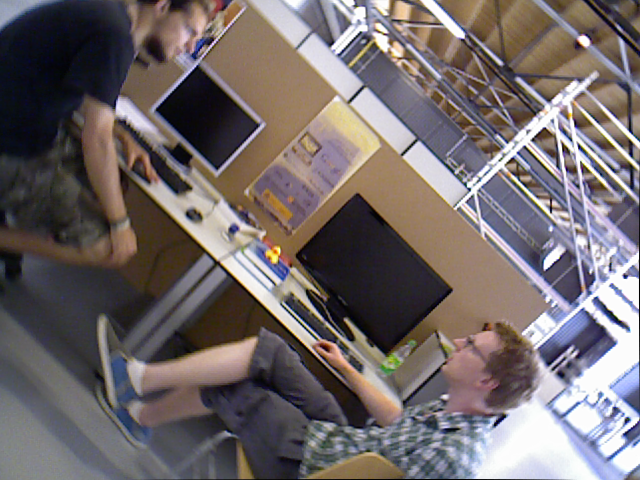}%
		\end{minipage}
	}
	\subfloat{%
		\begin{minipage}[b]{0.20\linewidth}
			\includegraphics[width=1\linewidth]{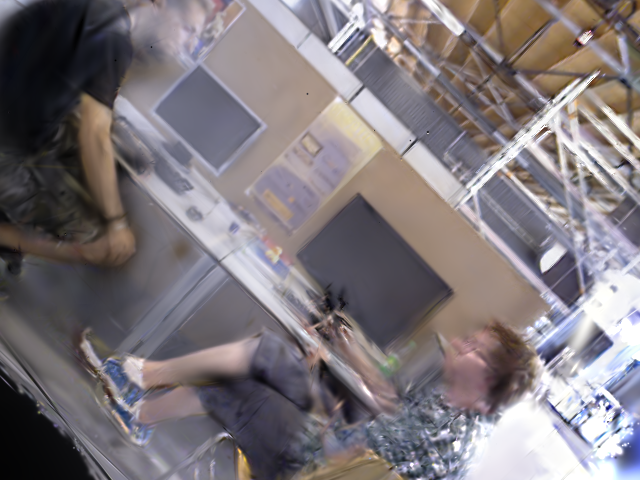}%
		\end{minipage}
	}
	\subfloat{%
		\begin{minipage}[b]{0.20\linewidth}
			\includegraphics[width=1\linewidth]{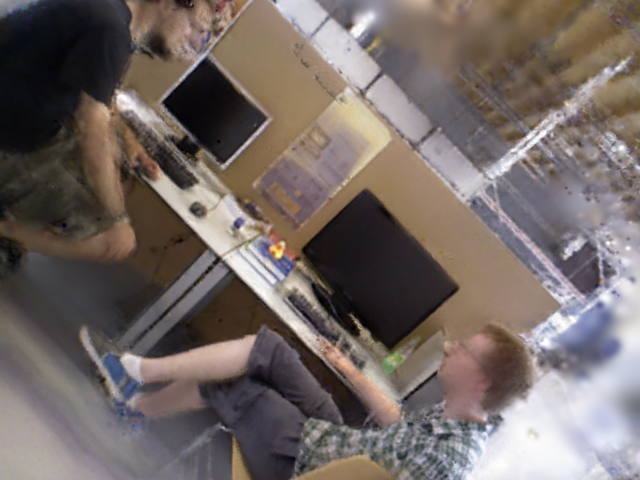}%
		\end{minipage}
	}
    \subfloat{%
		\begin{minipage}[b]{0.20\linewidth}
			\includegraphics[width=1\linewidth]{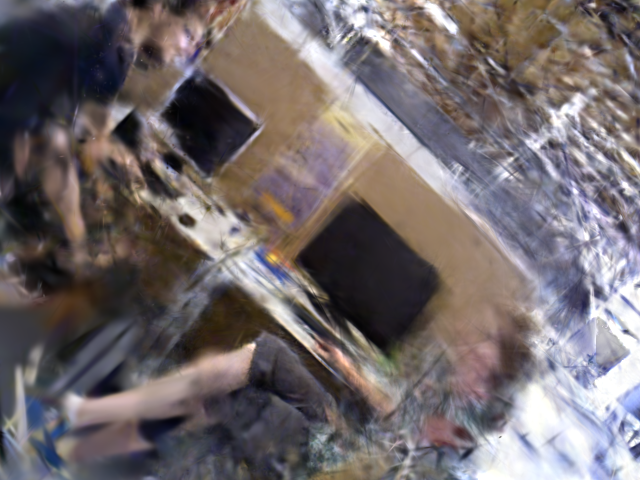}%
		\end{minipage}
	}
    \subfloat{%
		\begin{minipage}[b]{0.20\linewidth}
			\includegraphics[width=1\linewidth]{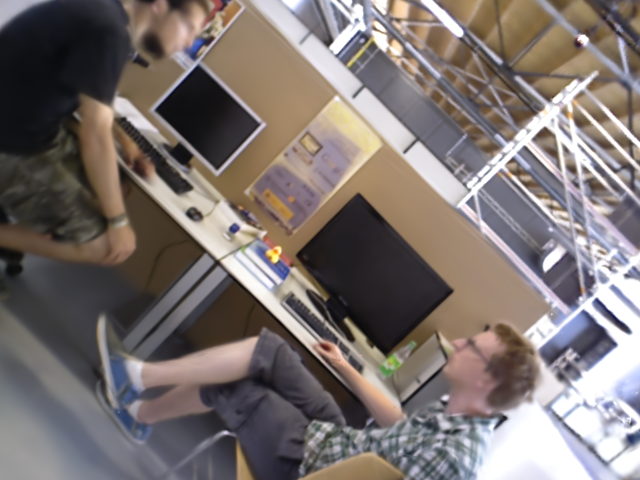}%
		\end{minipage}
	}\\
     % 第二行
	\vspace{1mm}
    \subfloat{%
        \hspace{-5mm}%
         \rotatebox{90}{\scriptsize{~~~~~~~~~\textbf{fr3\_sitting\_rpy}}}
		\begin{minipage}[b]{0.20\linewidth}
			\includegraphics[width=1\linewidth]{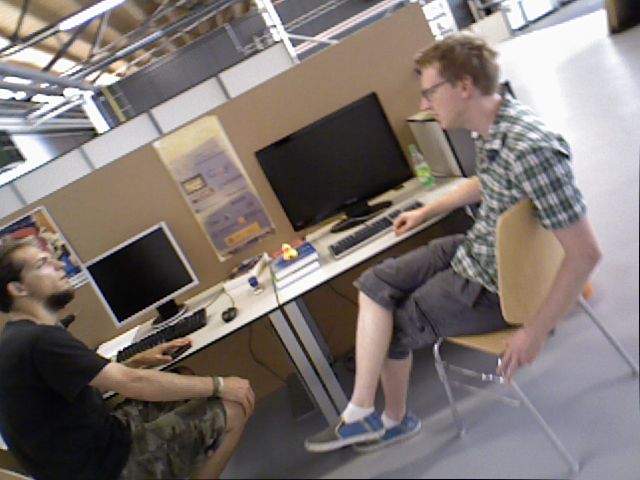}%
		\end{minipage}
	}
	\subfloat{%
		\begin{minipage}[b]{0.20\linewidth}
			\includegraphics[width=1\linewidth]{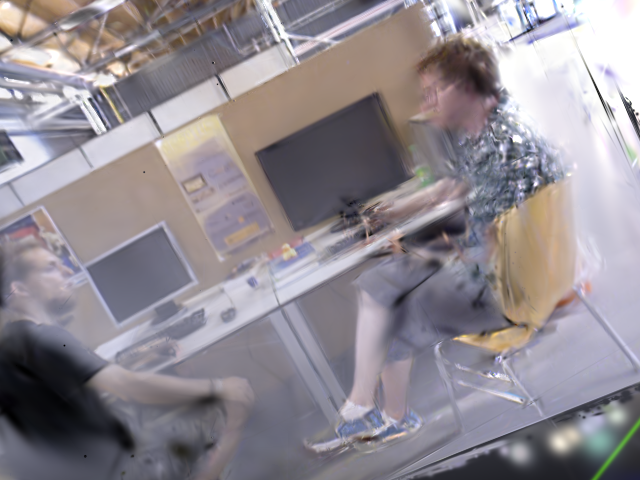}%
		\end{minipage}
	}
	\subfloat{%
		\begin{minipage}[b]{0.20\linewidth}
			\includegraphics[width=1\linewidth]{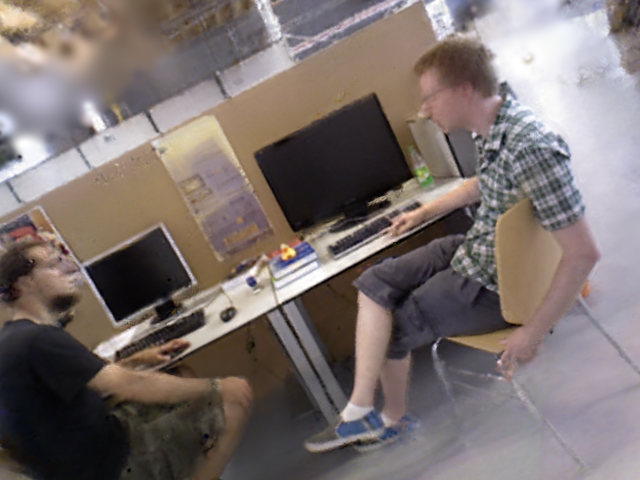}%
		\end{minipage}
	}
    \subfloat{%
		\begin{minipage}[b]{0.20\linewidth}
			\includegraphics[width=1\linewidth]{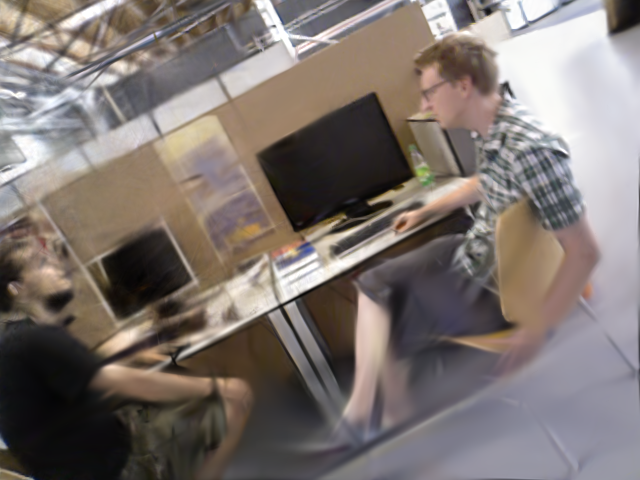}%
		\end{minipage}
	}
    \subfloat{%
		\begin{minipage}[b]{0.20\linewidth}
			\includegraphics[width=1\linewidth]{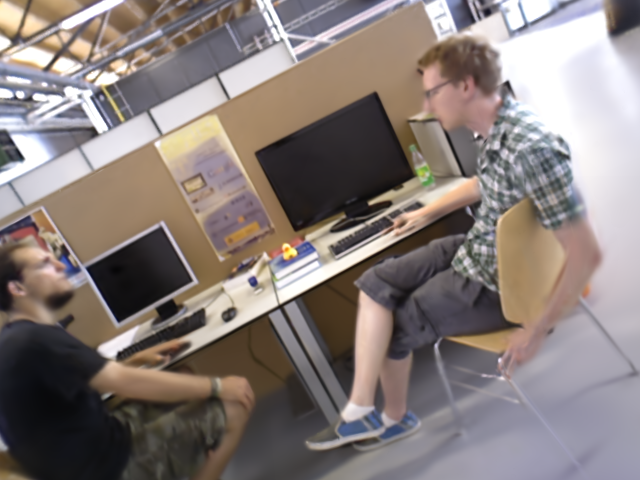}%
		\end{minipage}
	}\\
	\vspace{1mm}
    %第三行
    \subfloat[Ground Truth]{%
        \hspace{-5mm}%
        \rotatebox{90}{\scriptsize{~~~~~~\textbf{fr3\_walking\_static}}}
		\begin{minipage}[b]{0.20\linewidth}
			\includegraphics[width=1\linewidth]{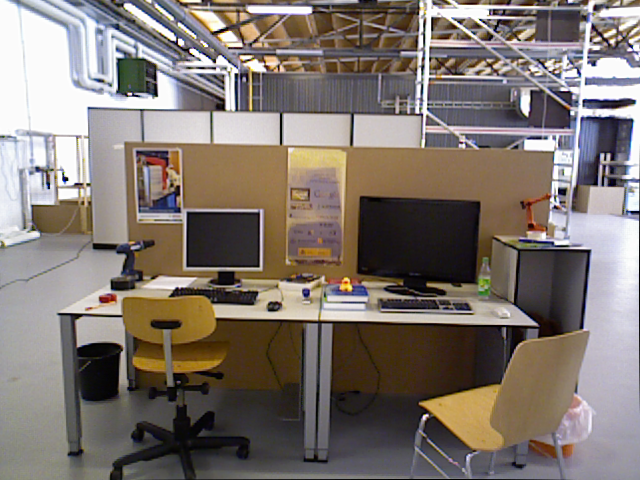}%
		\end{minipage}
	}
	\subfloat[MonoGS~\cite{Matsuki:Murai:etal:CVPR2024}]{%
		\begin{minipage}[b]{0.20\linewidth}
			\includegraphics[width=1\linewidth]{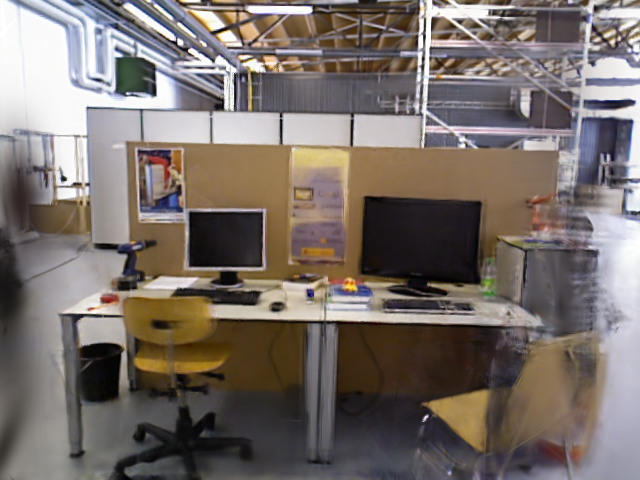}%
		\end{minipage}
	}
	\subfloat[SplaTAM~\cite{keetha2024splatam}]{%
		\begin{minipage}[b]{0.20\linewidth}
			\includegraphics[width=1\linewidth]{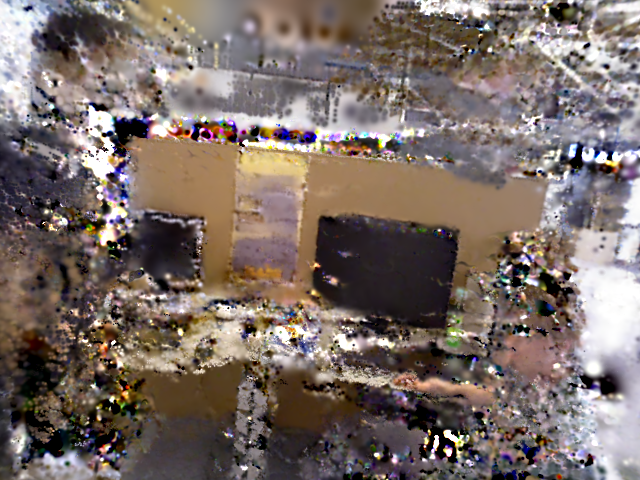}%
		\end{minipage}
	}
    \subfloat[SC-GS~\cite{huang2023sc}]{%
		\begin{minipage}[b]{0.20\linewidth}
			\includegraphics[width=1\linewidth]{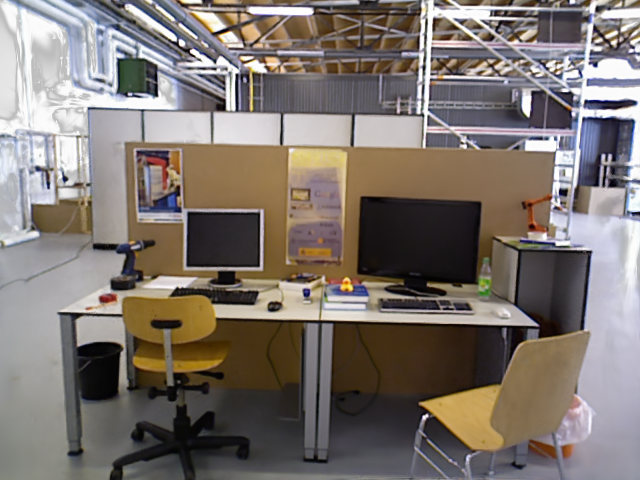}%
		\end{minipage}
	}
    \subfloat[Ours]{%
		\begin{minipage}[b]{0.20\linewidth}
			\includegraphics[width=1\linewidth]{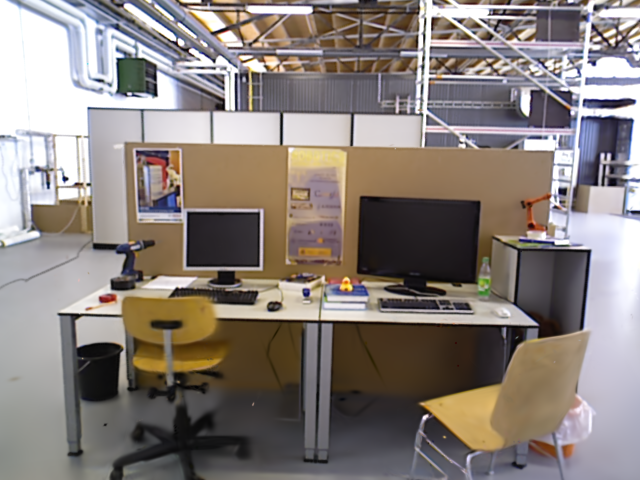}%
		\end{minipage}
	}\\

	\caption{\textbf{Visual comparison of the rendering images on the TUM RGB-D dataset.} }
	\label{TUM vis supp}
\end{figure*}

\paragraph{Color Refinement}
Finally, we perform 1500 iterations of global optimization. In each iteration, we randomly select 10 frames from all keyframes to optimize the dynamic deformation network and Gaussian parameters. The loss used is
\begin{equation}
    \begin{aligned}
        Loss =\ & 0.2 D\text{-}SSIM + 0.8 L_1(C(p)) \\
        &+ 0.1 L_1(D(p))+ W_1  \text{arap\_loss}+ W_2 E_{\text{iso}}\\
    \end{aligned}
\end{equation}
here, $W_1$ and $W_2$ are fixed weights.
% 对比实验1

\begin{figure}
	\centering
    \captionsetup[subfloat]{labelformat=empty}
        \subfloat[(a)GT]{%
        % \hspace{-5mm}%
		% \begin{minipage}[b]{0.5\linewidth}
			\includegraphics[width=0.48\linewidth]{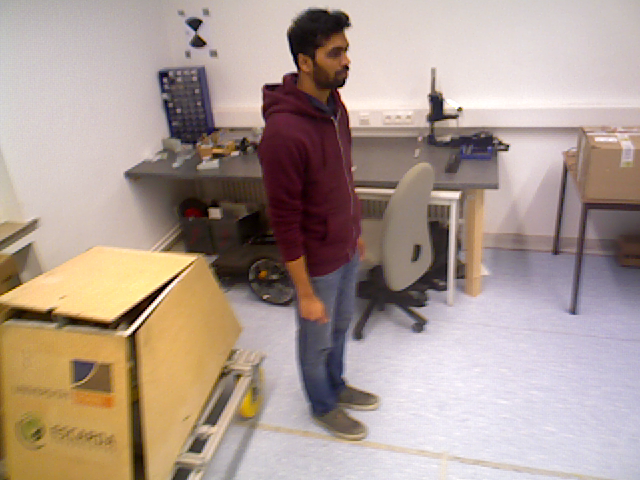}%
		% \end{minipage}
	}
	\subfloat[(b)8W2R]{\label{8w2r}
		% \begin{minipage}[b]{0.5\linewidth}
			\includegraphics[width=0.48\linewidth]{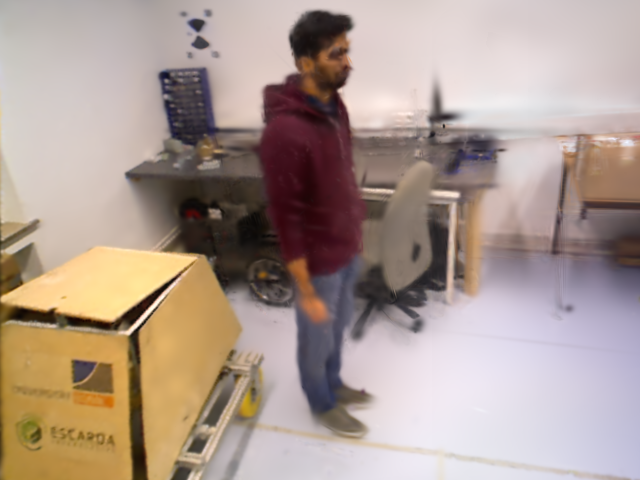}%
		% \end{minipage}
	}
       \newline
	% \vspace{-2mm}
	\subfloat[(c)5W5R]{\label{5w5r}
    \hspace{-0.9mm}%
		% \begin{minipage}[b]{0.5\linewidth}
			\includegraphics[width=0.48\linewidth]{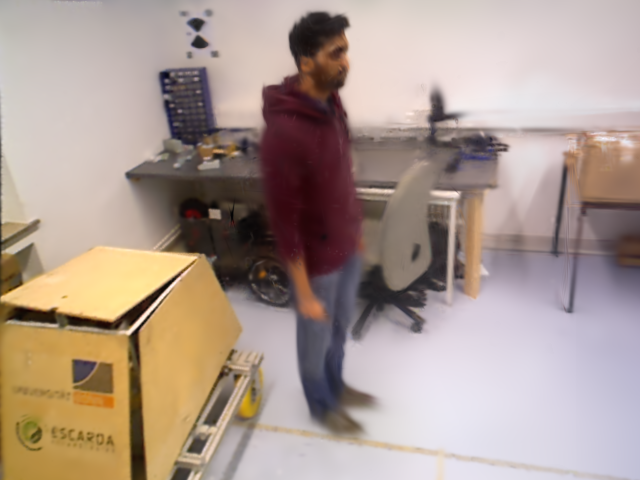}%
		% \end{minipage}
	}
        \subfloat[(d)1W7O2R]{\label{1w7o2r}
		% \begin{minipage}[b]{0.5\linewidth}
			\includegraphics[width=0.48\linewidth]{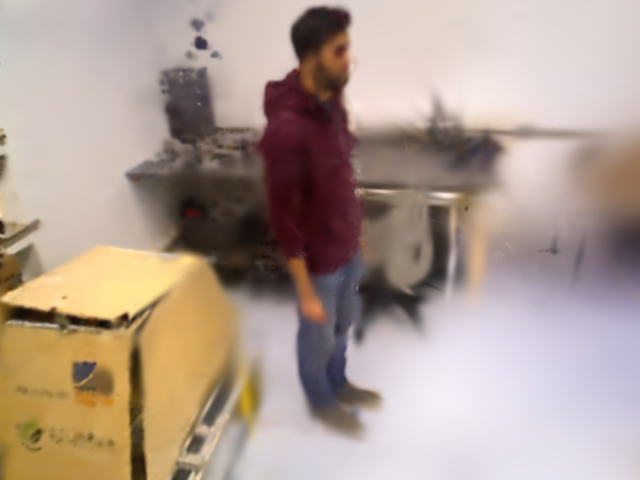}%
		% \end{minipage}
	}
	% \vspace{-2mm}
    \newline
    % \hspace{-1mm}%
        \subfloat[(e)w/o two-stage mapping]{\label{two-stage mapping}
     \hspace{-1.6mm}%
        % \hspace{-5mm}%
		% \begin{minipage}[b]{0.5\linewidth}
			\includegraphics[width=0.48\linewidth]{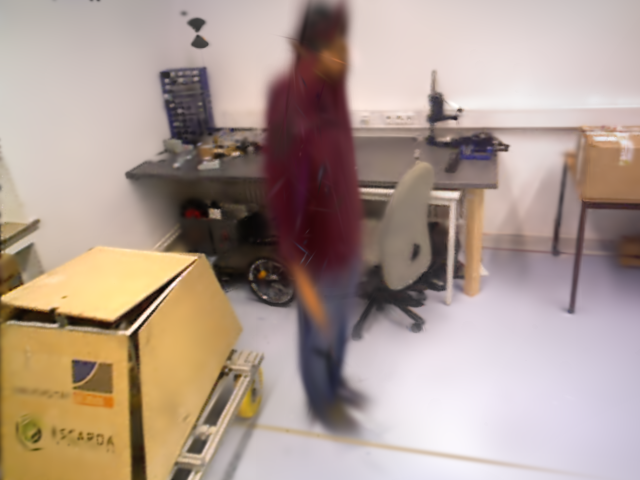}%
		% \end{minipage}
	}
	\subfloat[(f)fin]{\label{fin}
		% \begin{minipage}[b]{0.5\linewidth}
			\includegraphics[width=0.48\linewidth]{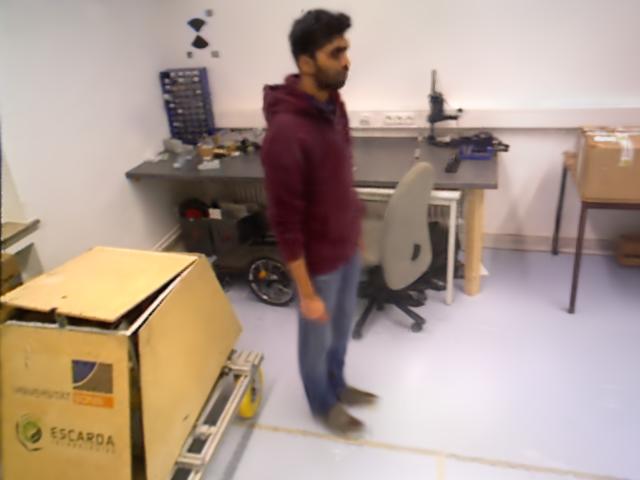}%
		% \end{minipage}
	}
	\caption{\textbf{The comparison of rendering results with different mapping strategies}. }
	\label{mapping strategies}
\end{figure}

% BONN数据集渲染结果对比
\begin{figure*}
	\centering
    \captionsetup[subfloat]{labelformat=empty}
	% \subfloat{%
 %        \hspace{-4.5mm}%
 %        \rotatebox{90}{\scriptsize{~~~~~~~~~~~~~~\textbf{ballon}}}
	% 	\begin{minipage}[b]{0.20\linewidth}
	% 		\includegraphics[width=1\linewidth]{imgs/study/ballon_0/gt.png}%
	% 	\end{minipage}
	% }
	% \subfloat{%
	% 	\begin{minipage}[b]{0.20\linewidth}
	% 		\includegraphics[width=1\linewidth]{imgs/study/ballon_0/monogs.png}%
	% 	\end{minipage}
	% }
	% \subfloat{%
	% 	\begin{minipage}[b]{0.20\linewidth}
	% 		\includegraphics[width=1\linewidth]{imgs/study/ballon_0/splatam.png}%
	% 	\end{minipage}
	% }
 %    \subfloat{%
	% 	\begin{minipage}[b]{0.20\linewidth}
	% 		\includegraphics[width=1\linewidth]{imgs/study/ballon_0/sc-gs.png}%
	% 	\end{minipage}
	% }
 %    \subfloat{%
	% 	\begin{minipage}[b]{0.20\linewidth}
	% 		\includegraphics[width=1\linewidth]{imgs/study/ballon_0/ours.png}%
	% 	\end{minipage}
	% }\\
 %     % 第二行
	 \vspace{1mm}
 %    \subfloat{%
 %        \hspace{-5mm}%
 %         \rotatebox{90}{\scriptsize{~~~~~~~~~~~\textbf{synchronous}}}
	% 	\begin{minipage}[b]{0.20\linewidth}
	% 		\includegraphics[width=1\linewidth]{imgs/study/syn/gt.png}%
	% 	\end{minipage}
	% }
	% \subfloat{%
	% 	\begin{minipage}[b]{0.20\linewidth}
	% 		\includegraphics[width=1\linewidth]{imgs/study/syn/MonoGS.png}%
	% 	\end{minipage}
	% }
	% \subfloat{%
	% 	\begin{minipage}[b]{0.20\linewidth}
	% 		\includegraphics[width=1\linewidth]{imgs/study/syn/splatam.png}%
	% 	\end{minipage}
	% }
 %    \subfloat{%
	% 	\begin{minipage}[b]{0.20\linewidth}
	% 		\includegraphics[width=1\linewidth]{imgs/study/syn/sc-gs.png}%
	% 	\end{minipage}
	% }
 %    \subfloat{%
	% 	\begin{minipage}[b]{0.20\linewidth}
	% 		\includegraphics[width=1\linewidth]{imgs/study/syn/ours.png}%
	% 	\end{minipage}
	% }\\
	% \vspace{1mm}
    % 第三行
    \subfloat{%
        \hspace{-5mm}%
        \rotatebox{90}{\scriptsize{~~~~~~~\textbf{placing\_no\_box3}}}
		\begin{minipage}[b]{0.20\linewidth}
			\includegraphics[width=1\linewidth]{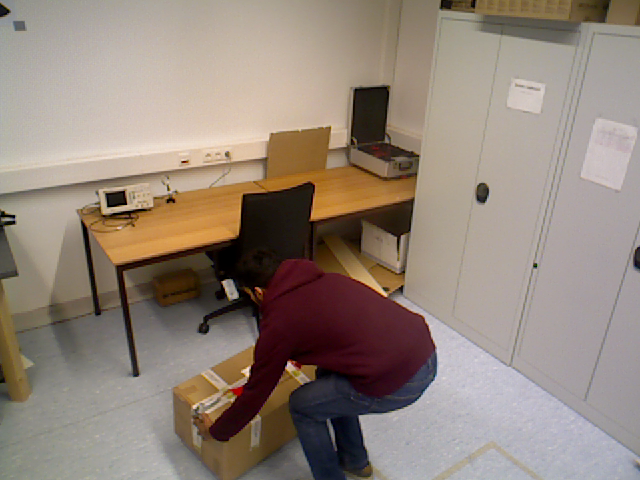}%
		\end{minipage}
	}
	\subfloat{%
		\begin{minipage}[b]{0.20\linewidth}
			\includegraphics[width=1\linewidth]{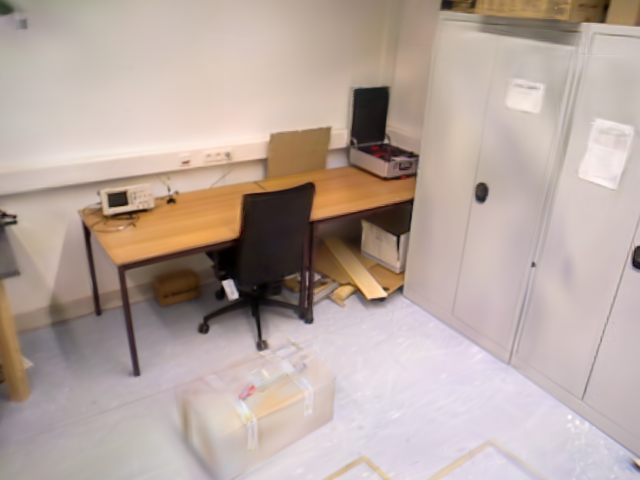}%
		\end{minipage}
	}
	\subfloat{%
		\begin{minipage}[b]{0.20\linewidth}
			\includegraphics[width=1\linewidth]{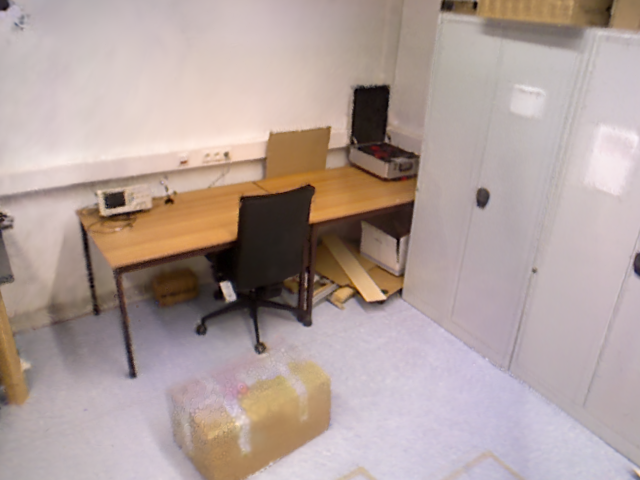}%
		\end{minipage}
	}
    \subfloat{%
		\begin{minipage}[b]{0.20\linewidth}
			\includegraphics[width=1\linewidth]{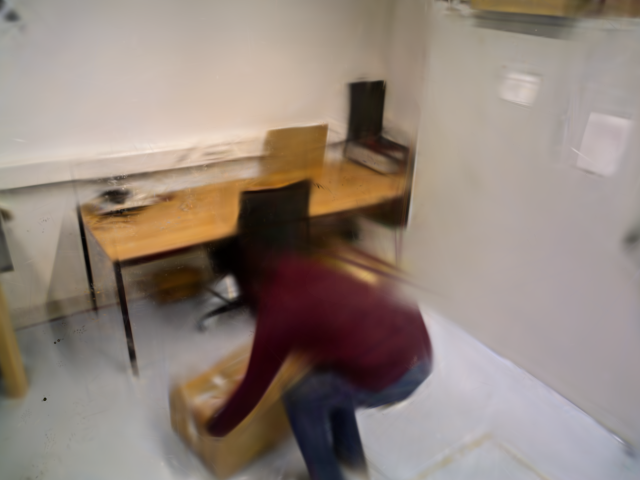}%
		\end{minipage}
	}
    \subfloat{%
		\begin{minipage}[b]{0.20\linewidth}
			\includegraphics[width=1\linewidth]{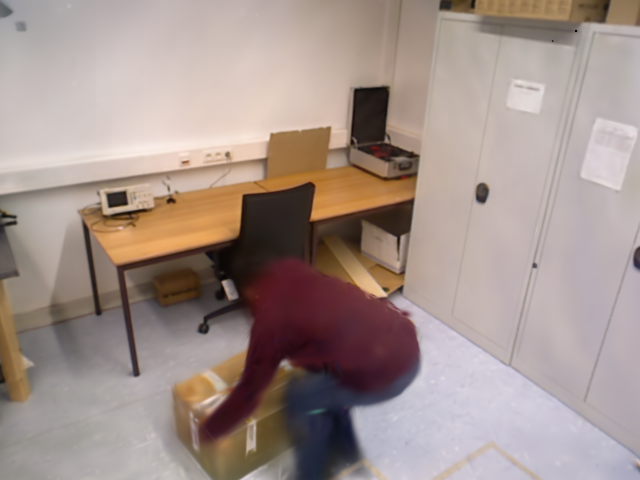}%
		\end{minipage}
	}\\
 %     \vspace{1mm}
 %    \subfloat{%
 %        \hspace{-5mm}%
 %        \rotatebox{90}{\scriptsize{~~~~~~~~~\textbf{synchronous2}}}
	% 	\begin{minipage}[b]{0.20\linewidth}
	% 		\includegraphics[width=1\linewidth]{imgs/study/syn2_250/gt.png}%
	% 	\end{minipage}
	% }
	% \subfloat{%
	% 	\begin{minipage}[b]{0.20\linewidth}
	% 		\includegraphics[width=1\linewidth]{imgs/study/syn2_250/monogs.png}%
	% 	\end{minipage}
	% }
	% \subfloat{%
	% 	\begin{minipage}[b]{0.20\linewidth}
	% 		\includegraphics[width=1\linewidth]{imgs/study/syn2_250/splatam.png}%
	% 	\end{minipage}
	% }
 %    \subfloat{%
	% 	\begin{minipage}[b]{0.20\linewidth}
	% 		\includegraphics[width=1\linewidth]{imgs/study/syn2_250/sc-gs.png}%
	% 	\end{minipage}
	% }
 %    \subfloat{%
	% 	\begin{minipage}[b]{0.20\linewidth}
	% 		\includegraphics[width=1\linewidth]{imgs/study/syn2_250/ours.png}%
	% 	\end{minipage}
	% }\\
     \vspace{1mm}
    \subfloat[Ground Truth]{%
        \hspace{-5mm}%
        \rotatebox{90}{\scriptsize{~~~~~~~~~~\textbf{synchronous}}}
		\begin{minipage}[b]{0.20\linewidth}
			\includegraphics[width=1\linewidth]{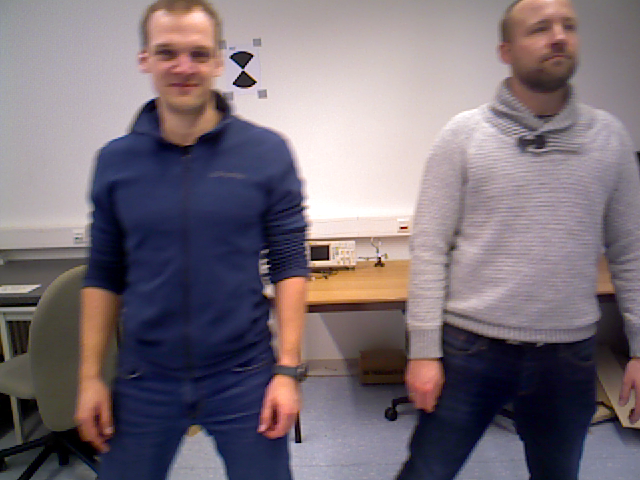}%
		\end{minipage}
	}
	\subfloat[MonoGS~\cite{Matsuki:Murai:etal:CVPR2024}]{%
		\begin{minipage}[b]{0.20\linewidth}
			\includegraphics[width=1\linewidth]{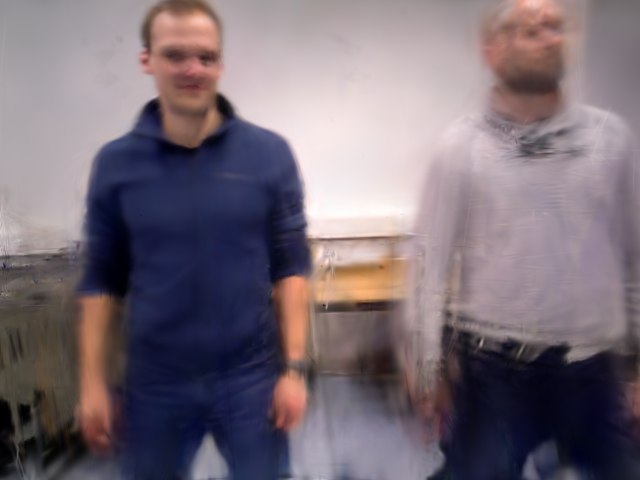}%
		\end{minipage}
	}
	\subfloat[SplaTAM~\cite{keetha2024splatam}]{%
		\begin{minipage}[b]{0.20\linewidth}
			\includegraphics[width=1\linewidth]{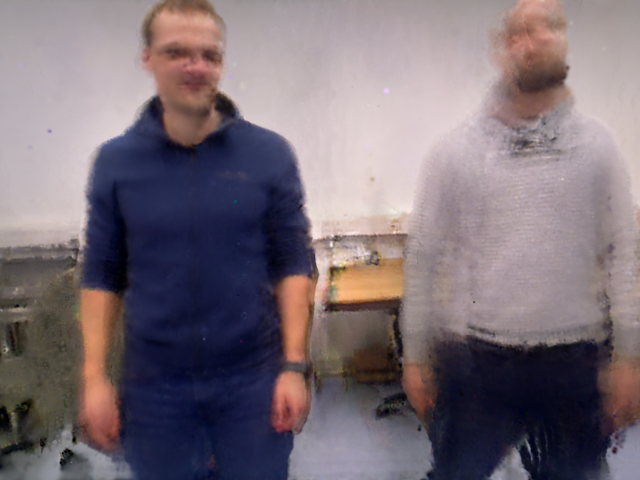}%
		\end{minipage}
	}
    \subfloat[SC-GS~\cite{huang2023sc}]{%
		\begin{minipage}[b]{0.20\linewidth}
			\includegraphics[width=1\linewidth]{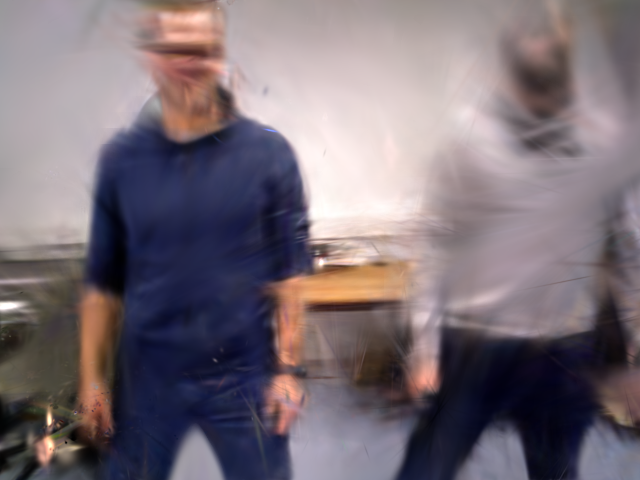}%
		\end{minipage}
	}
    \subfloat[Ours]{%
		\begin{minipage}[b]{0.20\linewidth}
			\includegraphics[width=1\linewidth]{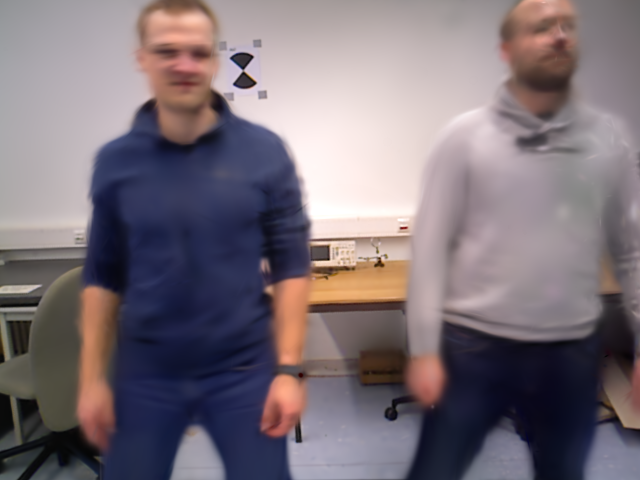}%
		\end{minipage}
	}\\

	\caption{\textbf{Visual comparison of the rendering image on the BONN RGB-D dataset.} This is also supported by the quantitative results in Table~\ref{BONN render}. More qualitative results have been added in Supplementary.}
	\label{BONN vis}
\end{figure*}

\section{Experiments}

\subsection{Dataset}
We evaluate our method on two real-world public datasets: the TUM RGB-D dataset\cite{sturm12iros} and the BONN RGB-D Dynamic dataset \cite{palazzolo2019iros}. Both datasets capture indoor scenes using a handheld camera and provide the ground-truth trajectories.

% \subsection{Dataset}
% We evaluate our method on two real-world public datasets: the TUM RGB-D dataset\cite{sturm12iros} and the BONN RGB-D Dynamic dataset \cite{palazzolo2019iros}. Both datasets capture indoor scenes using a handheld camera and provide the ground-truth trajectories.

\subsection{Implementation}
Our method is implemented in Python using the PyTorch framework, incorporating CUDA code for time-critical rasterization and gradient computation of Gaussian splatting, and we run our SLAM on a desktop with Intel(R) Xeon(R) Silver 4210R and a single NVIDIA GeForce RTX 3090 Ti. 
Furthermore, we set the weight $W_1=1e-4$, $W_2=10$, $\lambda=0.9$, $\lambda_{flow}=3$ , $\sigma=0.01$, for all datasets.  

For sequences where dynamic objects appear in the middle, such as sequence placing\_nonobstructing\_box of the BONN dataset, we pre-specify the initial frame for initializing dynamic Gaussians and control points.
\subsection{Baselines and Metrics.}
We primarily compare our method to existing GS-SLAM methods such as SplaTAM~\cite{keetha2024splatam}, Gaussian-SLAM~\cite{yugay2023gaussianslam} and MonoGS~\cite{Matsuki:Murai:etal:CVPR2024}, as well as Dynamic Gaussian Splatting methods like SC-GS~\cite{huang2023sc}, and the NeRF-SLAM method for dynamic scenes, RoDyn-SLAM~\cite{jiang2024rodynslam}. Additionally, for SC-GS, we select every 5th frame from the dataset as the training set, providing the real camera trajectory and the 3D model obtained by our method for training.

We use standard photometric rendering quality metrics to evaluate the quality of the map quality, including Peak Signal-to-Noise Ratio (PSNR), Structural Similarity Index Measure (SSIM), and Learned Perceptual Image Patch Similarity (LPIPS). Given that camera pose estimation performance is crucial for SLAM methods, we also report the Root Mean Square Error (RMSE) of the Absolute Trajectory Error (ATE) across all frames.

% % 对比实验2
\begin{table}
    \centering
    \renewcommand{\arraystretch}{1.2} %rows, default value is 1.0
    %\captionsetup[table*]{singlelinecheck=off}
    \setlength{\tabcolsep}{5pt}
    % \resizebox{\columnwidth}{!}{
    \begin{tabular}{cccc}
    \toprule
        Optical Flow & Separate Gaussians & syn & syn2\\
        \hline
        \raisebox{-0.5ex}{\XSolidBrush} & \raisebox{-0.5ex}{\XSolidBrush}  &18.37 & 22.11 \\
        %\hline
        \raisebox{-0.5ex}{\XSolidBrush}  & \raisebox{-0.5ex}{\Checkmark} &22.87 & 24.84 \\
        %\hline
        \raisebox{-0.5ex}{\Checkmark} & \raisebox{-0.5ex}{\XSolidBrush}  &17.40   &21.03  \\
        %\hline
        \raisebox{-0.5ex}{\Checkmark}  &\raisebox{-0.5ex}{\Checkmark}  &\textbf{23.25} &\textbf{25.42}   \\
    \bottomrule
    \end{tabular}%}
\caption{Analysis of the impact of Optical Flow Loss and Separate Gaussians on quantitative results (PSNR [dB] $\uparrow$) for the \textit{synchronous} and \textit{synchronous2} sequences in the BONN RGB-D dynamic dataset.}
\label{flow}
\end{table}

\subsection{Pose Estimation} \label{VO}
Besides rendering performance in appearance, we also evaluate the pose estimation performance of these methods. As shown in Table~\ref{TUM ATE} and~\ref{BONN ATE}, the estimated trajectories are compared to the ground truth ones. Thanks to the motion masks and separation of dynamic Gaussians, the proposed method shows robust and accurate camera pose estimation results in high-dynamic scenes compared to these GS-based SLAM methods. Furthermore, our method achieves more accurate results in most of the scenes compared to the NeRF-based dynamic SLAM, RoDyn-SLAM~\cite{jiang2024rodynslam}.

\subsection{Quality of Reconstructed Map}
Table \ref{TUM render} and \ref{BONN render} demonstrates the quality of the reconstructed map on the TUM RGB-D~\cite{sturm12iros} and BONN~\cite{palazzolo2019iros} datasets, respectively. We evaluated rendering quality by averaging the differences between the rendered images and the ground truth images across all frames. As shown in Figure~\ref{TUM render} and~\ref{BONN render}, our proposed method achieves better reconstruction than GS-based SLAM and dynamic Gaussian splatting SC-GS~\cite{huang2023sc} in most scenes. Due to the influence of exposure parameters, our method may perform slightly worse on some sequence metrics compared to other methods. However, as shown in \ref{BONN vis}, our method achieves the best reconstruction of static scenes and dynamic objects. More rendering results are provided in the supplementary material.

\subsection{Ablation Study}
\paragraph{Mapping Strategy}
In Figure \ref{mapping strategies}, we show the impact of different mapping strategies on the final rendering result. Figure~\ref{8w2r} represents the results of optimizing the first eight keyframes in the keyframe window and two randomly selected keyframes from all keyframes during the mapping process. Figure~\ref{5w5r} represents the results of optimizing the first five keyframes in the keyframe window and five randomly selected keyframes from all keyframes during the mapping process. Figure~\ref{1w7o2r} presents the results of optimizing the first keyframes in the keyframe window, two randomly selected keyframes from all keyframes, and seven randomly chosen keyframes that overlap with the current frame during the mapping process. Figure~\ref{two-stage mapping} shows the result of applying the same operation in the first-stage mapping as in the second-stage mapping, the keyframe selection during mapping is the same as in Figure~\ref{fin}. Figure~\ref{fin} is the method we use for mapping, which presents the results of optimizing the first three keyframes in the keyframe window, two randomly selected keyframes from all keyframes, and five randomly chosen keyframes that overlap with the current frame during the mapping process, achieving the best result in both dynamic and static scene reconstruction.

\paragraph{Optical-flow Loss and Separate Gaussians}
In Table~\ref{flow}, We ablate two aspects of our system: (1) whether optical flow loss is used during the mapping stage, and (2) whether only the dynamic Gaussian deformation is learned. We do this using \textit{synchronous} and \textit{synchronous2} sequences of the BONN dataset. The results presented in Table~\ref{flow} demonstrate the combined use of optical flow loss and dynamic Gaussian separation is effective in scene reconstruction.

\begin{figure*}
	\centering
%     \captionsetup[subfloat]{labelformat=empty}
	\subfloat{%
        \hspace{-5mm}%
        % \rotatebox{90}{\scriptsize{~~~~~~~\textbf{fr3\_sitting\_static}}}
		\begin{minipage}[b]{0.49\linewidth}
			\includegraphics[width=1\linewidth]{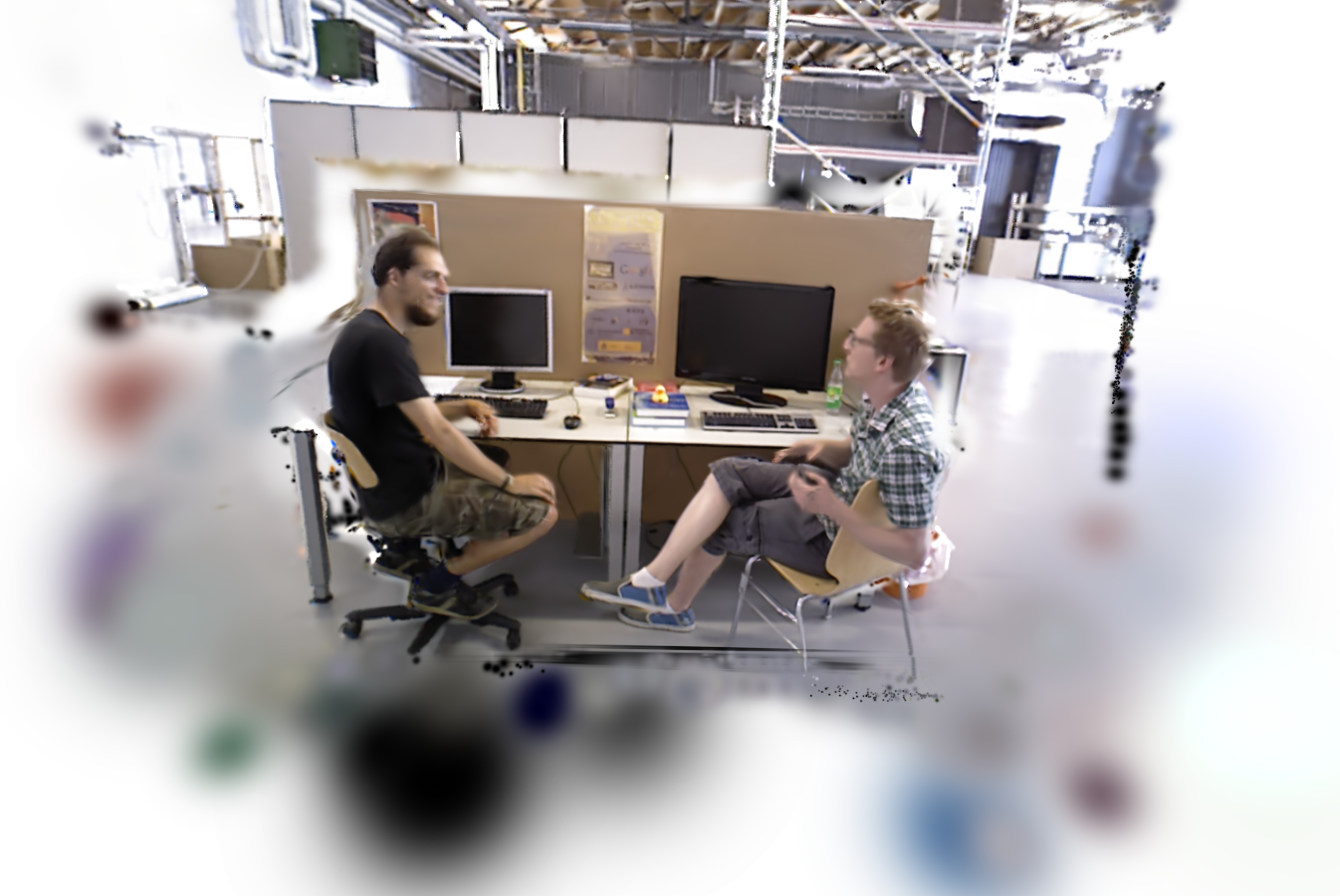}%
		\end{minipage}
	}
	\subfloat{%
		\begin{minipage}[b]{0.49\linewidth}
			\includegraphics[width=1\linewidth]{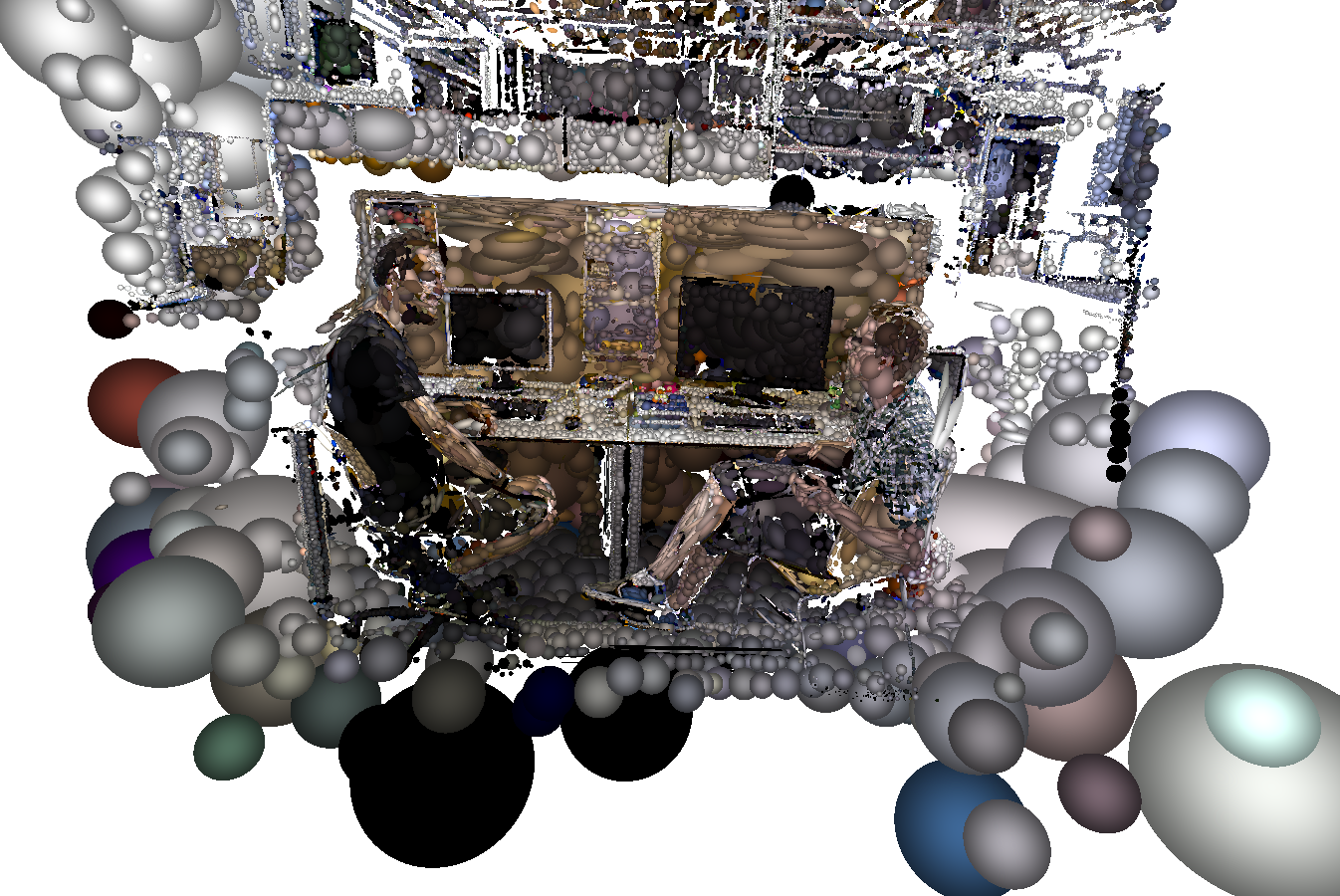}%
		\end{minipage}
	}
        \captionof{figure}{\small
        \textbf{Novel view rendering and Gaussian visualizations in the fr3\_sitting\_static sequence of TUM RGB-D.} }
 	% \label{BONN vis novel}
        \label{TUM vis novel}
    \end{figure*}

% TUM数据集渲染结果对比
\begin{figure*}
	\centering
    \captionsetup[subfloat]{labelformat=empty}
	\subfloat{%
        \hspace{-5mm}%
        \rotatebox{90}{\scriptsize{~~~~~~~\textbf{fr3\_sitting\_static}}}
		\begin{minipage}[b]{0.20\linewidth}
			\includegraphics[width=1\linewidth]{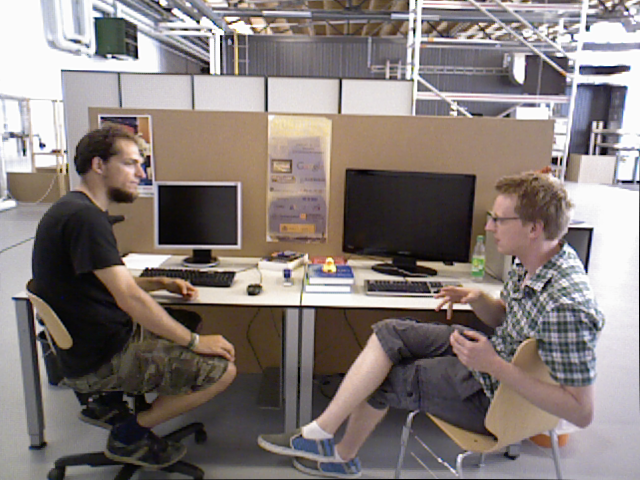}%
		\end{minipage}
	}
	\subfloat{%
		\begin{minipage}[b]{0.20\linewidth}
			\includegraphics[width=1\linewidth]{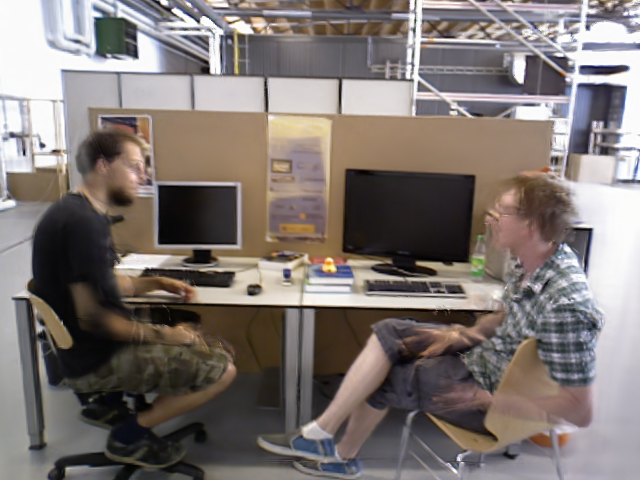}%
		\end{minipage}
	}
	\subfloat{%
		\begin{minipage}[b]{0.20\linewidth}
			\includegraphics[width=1\linewidth]{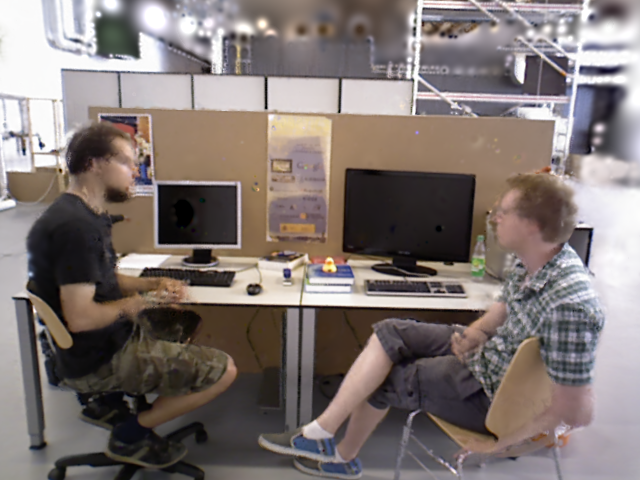}%
		\end{minipage}
	}
    \subfloat{%
		\begin{minipage}[b]{0.20\linewidth}
			\includegraphics[width=1\linewidth]{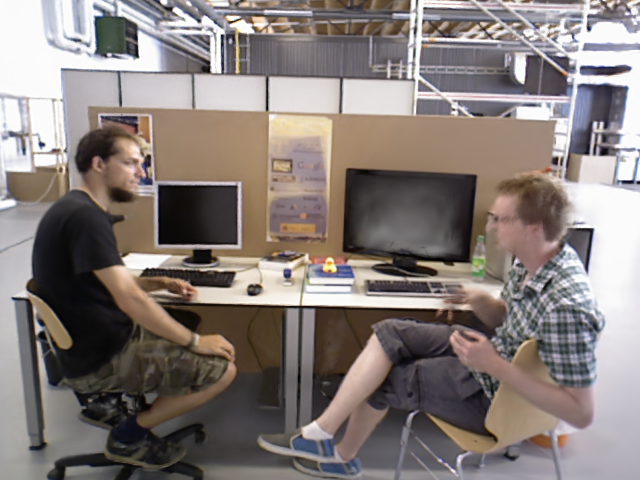}%
		\end{minipage}
	}
    \subfloat{%
		\begin{minipage}[b]{0.20\linewidth}
			\includegraphics[width=1\linewidth]{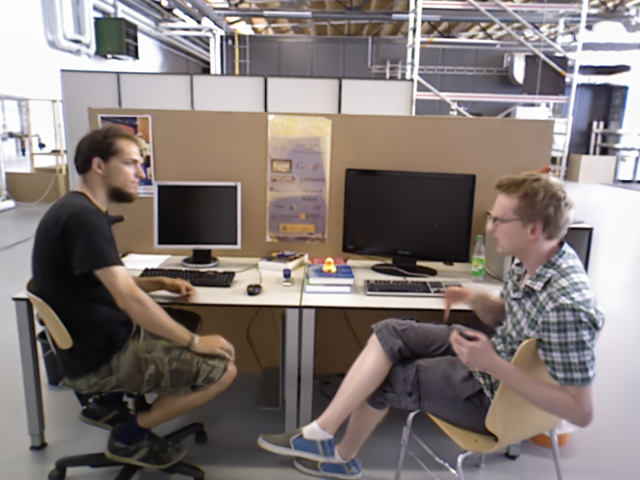}%
		\end{minipage}
	}\\
     % 第二行
	\vspace{1mm}
    \subfloat{%
        \hspace{-5mm}%
         \rotatebox{90}{\scriptsize{~~~~~~~\textbf{fr3\_sitting\_static}}}
		\begin{minipage}[b]{0.20\linewidth}
			\includegraphics[width=1\linewidth]{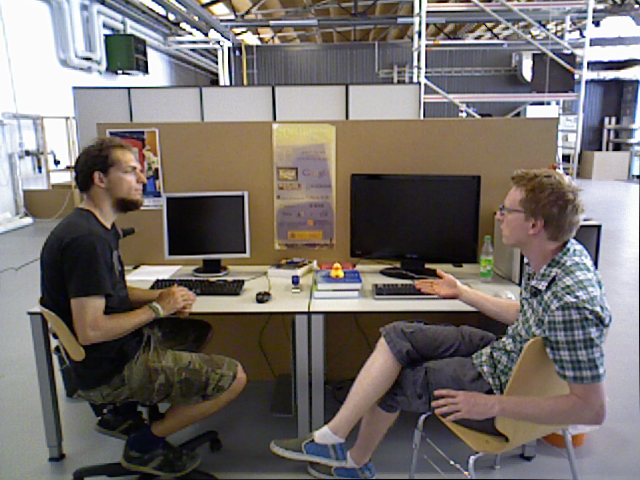}%
		\end{minipage}
	}
	\subfloat{%
		\begin{minipage}[b]{0.20\linewidth}
			\includegraphics[width=1\linewidth]{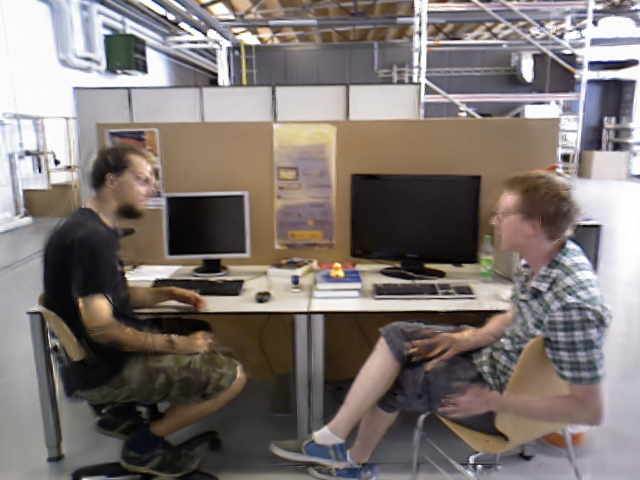}%
		\end{minipage}
	}
	\subfloat{%
		\begin{minipage}[b]{0.20\linewidth}
			\includegraphics[width=1\linewidth]{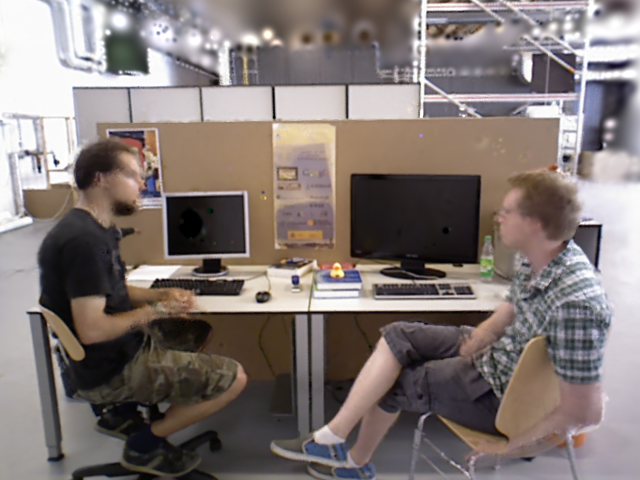}%
		\end{minipage}
	}
    \subfloat{%
		\begin{minipage}[b]{0.20\linewidth}
			\includegraphics[width=1\linewidth]{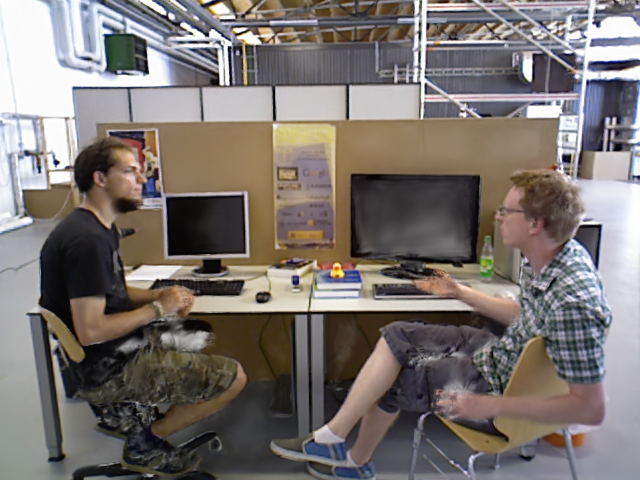}%
		\end{minipage}
	}
    \subfloat{%
		\begin{minipage}[b]{0.20\linewidth}
			\includegraphics[width=1\linewidth]{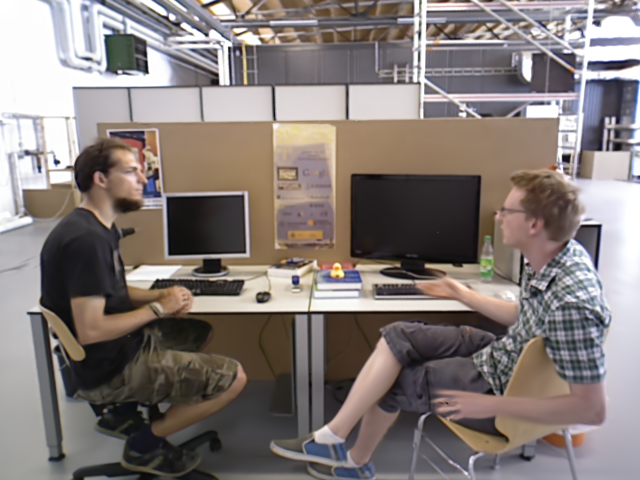}%
		\end{minipage}
	}\\
	\vspace{1mm}
    % 第三行
    \subfloat{%
        \hspace{-5mm}%
        \rotatebox{90}{\scriptsize{~~~~~~~~~\textbf{fr3\_sitting\_xyz}}}
		\begin{minipage}[b]{0.20\linewidth}
			\includegraphics[width=1\linewidth]{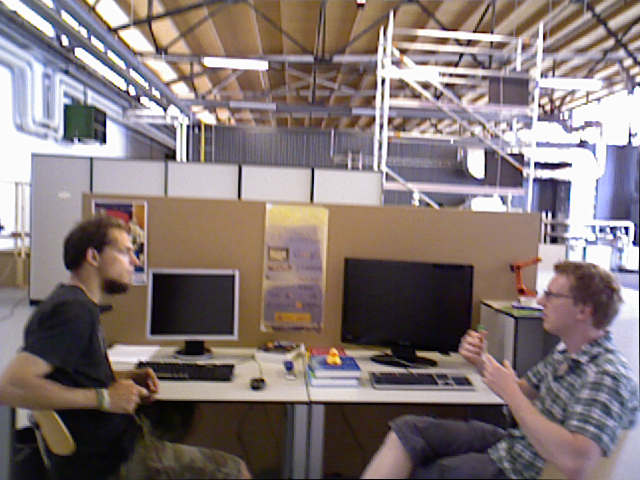}%
		\end{minipage}
	}
	\subfloat{%
		\begin{minipage}[b]{0.20\linewidth}
			\includegraphics[width=1\linewidth]{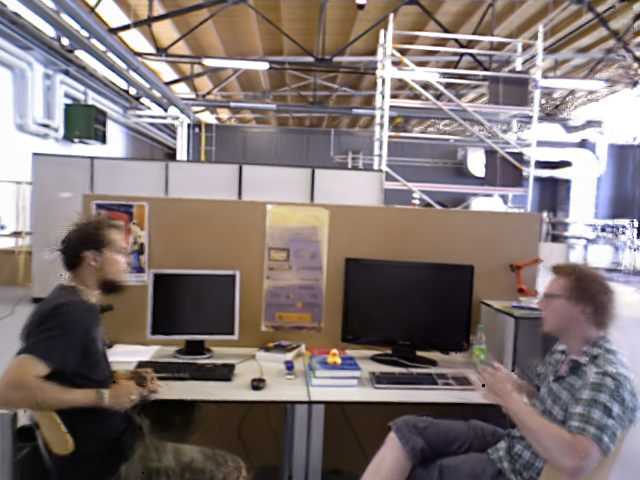}%
		\end{minipage}
	}
	\subfloat{%
		\begin{minipage}[b]{0.20\linewidth}
			\includegraphics[width=1\linewidth]{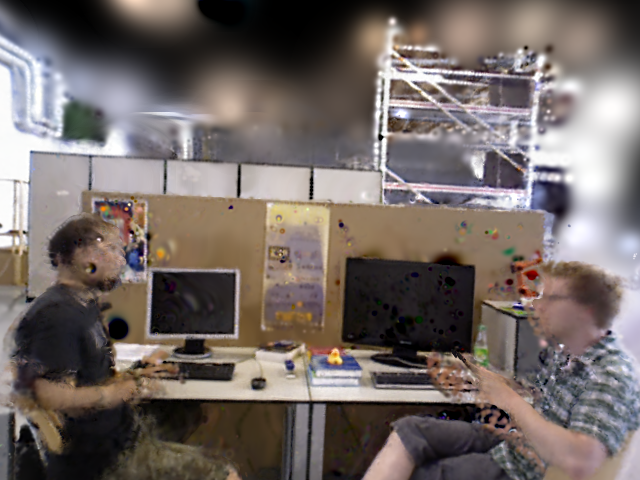}%
		\end{minipage}
	}
    \subfloat{%
		\begin{minipage}[b]{0.20\linewidth}
			\includegraphics[width=1\linewidth]{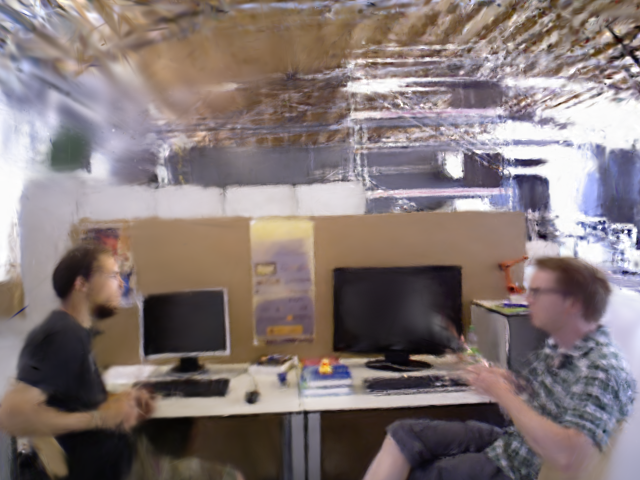}%
		\end{minipage}
	}
    \subfloat{%
		\begin{minipage}[b]{0.20\linewidth}
			\includegraphics[width=1\linewidth]{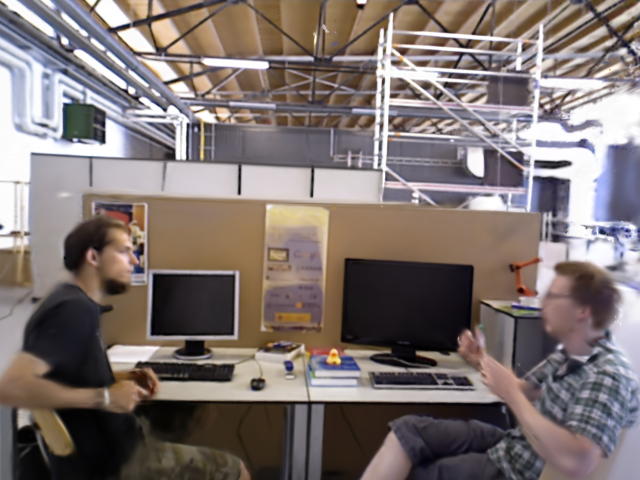}%
		\end{minipage}
	}\\
     \vspace{1mm}
    %第四行
    \subfloat[Ground Truth]{%
        \hspace{-5mm}%
        \rotatebox{90}{\scriptsize{~~~~~~~~~\textbf{fr3\_sitting\_xyz}}}
		\begin{minipage}[b]{0.20\linewidth}
			\includegraphics[width=1\linewidth]{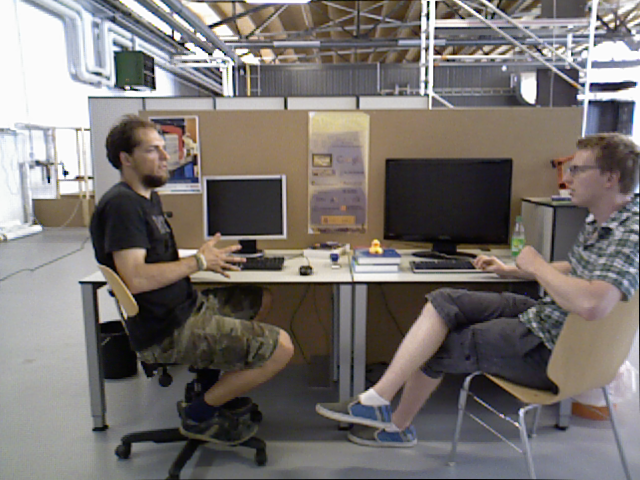}%
		\end{minipage}
	}
	\subfloat[MonoGS~\cite{Matsuki:Murai:etal:CVPR2024}]{%
		\begin{minipage}[b]{0.20\linewidth}
			\includegraphics[width=1\linewidth]{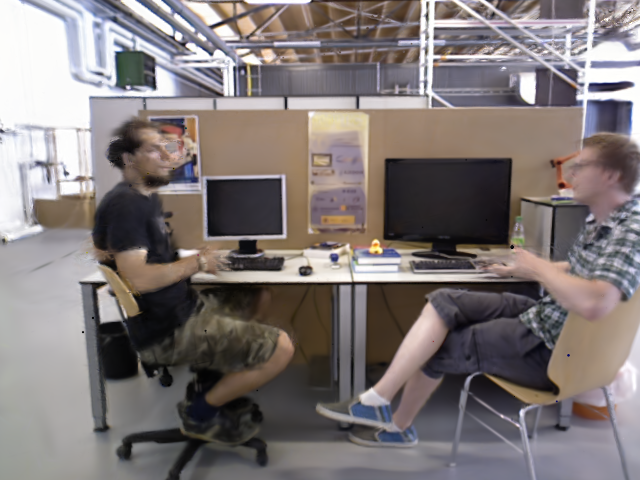}%
		\end{minipage}
	}
	\subfloat[SplaTAM~\cite{keetha2024splatam}]{%
		\begin{minipage}[b]{0.20\linewidth}
			\includegraphics[width=1\linewidth]{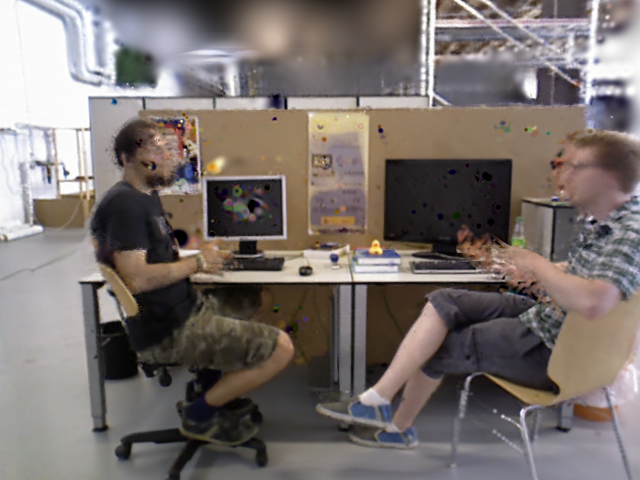}%
		\end{minipage}
	}
    \subfloat[SC-GS~\cite{huang2023sc}]{%
		\begin{minipage}[b]{0.20\linewidth}
			\includegraphics[width=1\linewidth]{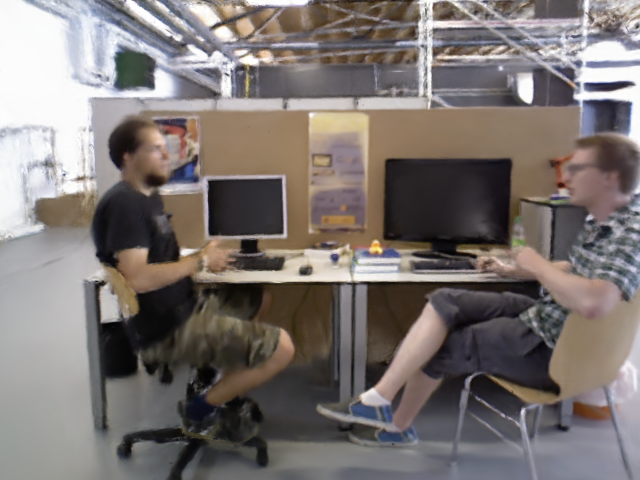}%
		\end{minipage}
	}
    \subfloat[Ours]{%
		\begin{minipage}[b]{0.20\linewidth}
			\includegraphics[width=1\linewidth]{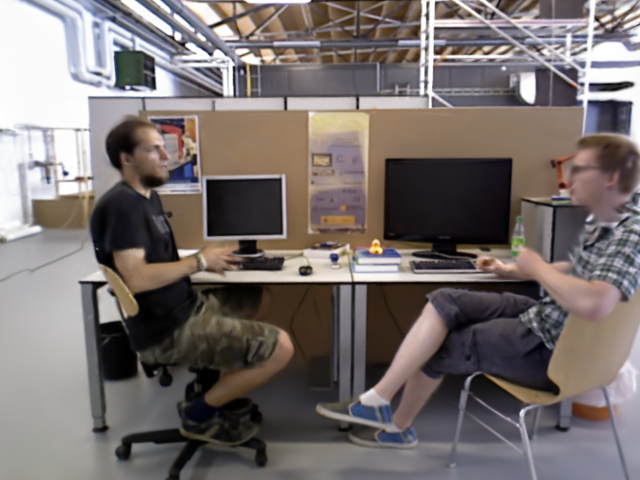}%
		\end{minipage}
	}\\

	\caption{\textbf{Visual comparison of the rendering image on the TUM RGB-D dataset.} }
	\label{TUM vis}
\end{figure*}

\section{Conclusion}
In this paper, we propose a novel approach for reconstructing dynamic scenes using 4D Gaussian Splatting SLAM. Our method incrementally tracks camera poses and reconstructs dynamic scenes from a sequence of RGB-D images in unknown environments. By leveraging the power of dynamic and static Gaussian segmentation and optical flow, our approach not only localizes the camera and reconstructs the static environment but also effectively maps dynamic objects. We demonstrate its effectiveness in achieving state-of-the-art results in camera pose estimation and dynamic scene reconstruction.

%% file: main.bbl
\begin{thebibliography}{54}
\providecommand{\natexlab}[1]{#1}
\providecommand{\url}[1]{\texttt{#1}}
\expandafter\ifx\csname urlstyle\endcsname\relax
  \providecommand{\doi}[1]{doi: #1}\else
  \providecommand{\doi}{doi: \begingroup \urlstyle{rm}\Url}\fi

\bibitem[Bae et~al.(2024)Bae, Kim, Yun, Lee, Bang, and Uh]{bae2024ed3dgs}
Jeongmin Bae, Seoha Kim, Youngsik Yun, Hahyun Lee, Gun Bang, and Youngjung Uh.
\newblock Per-gaussian embedding-based deformation for deformable 3d gaussian splatting.
\newblock In \emph{European Conference on Computer Vision (ECCV)}, 2024.

\bibitem[Barron et~al.(2022)Barron, Mildenhall, Verbin, Srinivasan, and Hedman]{barron2022mip}
Jonathan~T Barron, Ben Mildenhall, Dor Verbin, Pratul~P Srinivasan, and Peter Hedman.
\newblock Mip-nerf 360: Unbounded anti-aliased neural radiance fields.
\newblock In \emph{Proceedings of the IEEE/CVF Conference on Computer Vision and Pattern Recognition}, pages 5470--5479, 2022.

\bibitem[Campos et~al.(2021)Campos, Elvira, Rodr{\'\i}guez, Montiel, and Tard{\'o}s]{campos2021orb}
Carlos Campos, Richard Elvira, Juan J~G{\'o}mez Rodr{\'\i}guez, Jos{\'e}~MM Montiel, and Juan~D Tard{\'o}s.
\newblock Orb-slam3: An accurate open-source library for visual, visual--inertial, and multimap slam.
\newblock \emph{IEEE Transactions on Robotics}, 37\penalty0 (6):\penalty0 1874--1890, 2021.

\bibitem[Cao and Johnson(2023)]{Cao2023HexPlane}
Ang Cao and Justin Johnson.
\newblock Hexplane: A fast representation for dynamic scenes.
\newblock \emph{CVPR}, 2023.

\bibitem[Dai et~al.(2017)Dai, Nie{\ss}ner, Zollh{\"o}fer, Izadi, and Theobalt]{dai2017bundlefusion}
Angela Dai, Matthias Nie{\ss}ner, Michael Zollh{\"o}fer, Shahram Izadi, and Christian Theobalt.
\newblock Bundlefusion: Real-time globally consistent 3d reconstruction using on-the-fly surface reintegration.
\newblock \emph{ACM Transactions on Graphics (ToG)}, 36\penalty0 (4):\penalty0 1, 2017.

\bibitem[DeTone et~al.(2018)DeTone, Malisiewicz, and Rabinovich]{detone2018superpoint}
Daniel DeTone, Tomasz Malisiewicz, and Andrew Rabinovich.
\newblock Superpoint: Self-supervised interest point detection and description.
\newblock In \emph{Proceedings of the IEEE conference on computer vision and pattern recognition workshops}, pages 224--236, 2018.

\bibitem[Engel et~al.(2014)Engel, Sch{\"o}ps, and Cremers]{engel2014lsd}
Jakob Engel, Thomas Sch{\"o}ps, and Daniel Cremers.
\newblock Lsd-slam: Large-scale direct monocular slam.
\newblock In \emph{European conference on computer vision}, pages 834--849. Springer, 2014.

\bibitem[Engel et~al.(2015)Engel, St{\"u}ckler, and Cremers]{engel2015large}
Jakob Engel, J{\"o}rg St{\"u}ckler, and Daniel Cremers.
\newblock Large-scale direct slam with stereo cameras.
\newblock In \emph{2015 IEEE/RSJ international conference on intelligent robots and systems (IROS)}, pages 1935--1942. IEEE, 2015.

\bibitem[Fleet and Weiss(2006)]{fleet2006optical}
David Fleet and Yair Weiss.
\newblock Optical flow estimation.
\newblock In \emph{Handbook of mathematical models in computer vision}, pages 237--257. Springer, 2006.

\bibitem[Fu et~al.(2023)Fu, Liu, Kulkarni, Kautz, Efros, and Wang]{fu2023colmap}
Yang Fu, Sifei Liu, Amey Kulkarni, Jan Kautz, Alexei~A Efros, and Xiaolong Wang.
\newblock Colmap-free 3d gaussian splatting.
\newblock \emph{arXiv preprint arXiv:2312.07504}, 2023.

\bibitem[Ha et~al.(2024)Ha, Yeon, and Yu]{ha2024rgbdgsicpslam}
Seongbo Ha, Jiung Yeon, and Hyeonwoo Yu.
\newblock Rgbd gs-icp slam, 2024.

\bibitem[Haralick et~al.(1989)Haralick, Joo, Lee, Zhuang, Vaidya, and Kim]{haralick1989pose}
Robert~M Haralick, Hyonam Joo, Chung-Nan Lee, Xinhua Zhuang, Vinay~G Vaidya, and Man~Bae Kim.
\newblock Pose estimation from corresponding point data.
\newblock \emph{IEEE Transactions on Systems, Man, and Cybernetics}, 19\penalty0 (6):\penalty0 1426--1446, 1989.

\bibitem[Henein et~al.(2020)Henein, Zhang, Mahony, and Ila]{henein2020dynamic}
Mina Henein, Jun Zhang, Robert Mahony, and Viorela Ila.
\newblock Dynamic slam: The need for speed.
\newblock In \emph{2020 IEEE International Conference on Robotics and Automation (ICRA)}, pages 2123--2129. IEEE, 2020.

\bibitem[Hesch and Roumeliotis(2011)]{hesch2011direct}
Joel~A Hesch and Stergios~I Roumeliotis.
\newblock A direct least-squares (dls) method for pnp.
\newblock In \emph{2011 International Conference on Computer Vision}, pages 383--390. IEEE, 2011.

\bibitem[Hou et~al.(2025)Hou, Yeo, Guo, Su, Li, and Lee]{hou2025mvgsr}
Chenfeng Hou, Qi~Xun Yeo, Mengqi Guo, Yongxin Su, Yanyan Li, and Gim~Hee Lee.
\newblock Mvgsr: Multi-view consistency gaussian splatting for robust surface reconstruction.
\newblock \emph{arXiv preprint arXiv:2503.08093}, 2025.

\bibitem[Hou et~al.(2019)Hou, Dai, and Nie{\ss}ner]{hou20193d}
Ji Hou, Angela Dai, and Matthias Nie{\ss}ner.
\newblock 3d-sis: 3d semantic instance segmentation of rgb-d scans.
\newblock In \emph{Proceedings of the IEEE/CVF conference on computer vision and pattern recognition}, pages 4421--4430, 2019.

\bibitem[Hu et~al.(2024)Hu, Li, Xie, Xu, Dong, Yao, and Lee]{hu2024learnable}
Bingbing Hu, Yanyan Li, Rui Xie, Bo Xu, Haoye Dong, Junfeng Yao, and Gim~Hee Lee.
\newblock Learnable infinite taylor gaussian for dynamic view rendering.
\newblock \emph{arXiv preprint arXiv:2412.04282}, 2024.

\bibitem[Huang et~al.(2023)Huang, Sun, Yang, Lyu, Cao, and Qi]{huang2023sc}
Yi-Hua Huang, Yang-Tian Sun, Ziyi Yang, Xiaoyang Lyu, Yan-Pei Cao, and Xiaojuan Qi.
\newblock Sc-gs: Sparse-controlled gaussian splatting for editable dynamic scenes.
\newblock \emph{arXiv preprint arXiv:2312.14937}, 2023.

\bibitem[Izadi et~al.(2011)Izadi, Kim, Hilliges, Molyneaux, Newcombe, Kohli, Shotton, Hodges, Freeman, Davison, et~al.]{izadi2011kinectfusion}
Shahram Izadi, David Kim, Otmar Hilliges, David Molyneaux, Richard Newcombe, Pushmeet Kohli, Jamie Shotton, Steve Hodges, Dustin Freeman, Andrew Davison, et~al.
\newblock Kinectfusion: real-time 3d reconstruction and interaction using a moving depth camera.
\newblock In \emph{Proceedings of the 24th annual ACM symposium on User interface software and technology}, pages 559--568, 2011.

\bibitem[Jiang et~al.(2024)Jiang, Xu, Li, Feng, and Zhang]{jiang2024rodynslam}
Haochen Jiang, Yueming Xu, Kejie Li, Jianfeng Feng, and Li Zhang.
\newblock Rodyn-slam: Robust dynamic dense rgb-d slam with neural radiance fields.
\newblock \emph{IEEE Robotics and Automation Letters}, 2024.

\bibitem[Keetha et~al.(2024)Keetha, Karhade, Jatavallabhula, Yang, Scherer, Ramanan, and Luiten]{keetha2024splatam}
Nikhil Keetha, Jay Karhade, Krishna~Murthy Jatavallabhula, Gengshan Yang, Sebastian Scherer, Deva Ramanan, and Jonathon Luiten.
\newblock Splatam: Splat, track and map 3d gaussians for dense rgb-d slam.
\newblock In \emph{Proceedings of the IEEE/CVF Conference on Computer Vision and Pattern Recognition}, 2024.

\bibitem[Kerbl et~al.(2023)Kerbl, Kopanas, Leimk{\"u}hler, and Drettakis]{kerbl3Dgaussians}
Bernhard Kerbl, Georgios Kopanas, Thomas Leimk{\"u}hler, and George Drettakis.
\newblock 3d gaussian splatting for real-time radiance field rendering.
\newblock \emph{ACM Transactions on Graphics}, 42\penalty0 (4), 2023.

\bibitem[Kirillov et~al.(2023)Kirillov, Mintun, Ravi, Mao, Rolland, Gustafson, Xiao, Whitehead, Berg, Lo, et~al.]{kirillov2023segment}
Alexander Kirillov, Eric Mintun, Nikhila Ravi, Hanzi Mao, Chloe Rolland, Laura Gustafson, Tete Xiao, Spencer Whitehead, Alexander~C Berg, Wan-Yen Lo, et~al.
\newblock Segment anything.
\newblock In \emph{Proceedings of the IEEE/CVF international conference on computer vision}, pages 4015--4026, 2023.

\bibitem[Kong et~al.(2024)Kong, Lee, Lee, and Kim]{kong2024dgs}
Mangyu Kong, Jaewon Lee, Seongwon Lee, and Euntai Kim.
\newblock Dgs-slam: Gaussian splatting slam in dynamic environment.
\newblock \emph{arXiv preprint arXiv:2411.10722}, 2024.

\bibitem[Li et~al.(2020)Li, Brasch, Wang, Navab, and Tombari]{li2020structure}
Yanyan Li, Nikolas Brasch, Yida Wang, Nassir Navab, and Federico Tombari.
\newblock Structure-slam: Low-drift monocular slam in indoor environments.
\newblock \emph{IEEE Robotics and Automation Letters}, 5\penalty0 (4):\penalty0 6583--6590, 2020.

\bibitem[Li et~al.(2021)Li, Yunus, Brasch, Navab, and Tombari]{li2021rgb}
Yanyan Li, Raza Yunus, Nikolas Brasch, Nassir Navab, and Federico Tombari.
\newblock Rgb-d slam with structural regularities.
\newblock In \emph{2021 IEEE international conference on Robotics and automation (ICRA)}, pages 11581--11587. IEEE, 2021.

\bibitem[Li et~al.(2024)Li, Lyu, Di, Zhai, Lee, and Tombari]{li2024geogaussian}
Yanyan Li, Chenyu Lyu, Yan Di, Guangyao Zhai, Gim~Hee Lee, and Federico Tombari.
\newblock Geogaussian: Geometry-aware gaussian splatting for scene rendering.
\newblock In \emph{European Conference on Computer Vision}, pages 441--457. Springer, 2024.

\bibitem[Luong and Faugeras(1997)]{luong1997self}
Q-T Luong and Olivier~D Faugeras.
\newblock Self-calibration of a moving camera from point correspondences and fundamental matrices.
\newblock \emph{International Journal of computer vision}, 22:\penalty0 261--289, 1997.

\bibitem[Matsuki et~al.(2024{\natexlab{a}})Matsuki, Murai, Kelly, and Davison]{matsuki2024gaussian}
Hidenobu Matsuki, Riku Murai, Paul~HJ Kelly, and Andrew~J Davison.
\newblock Gaussian splatting slam.
\newblock In \emph{Proceedings of the IEEE/CVF Conference on Computer Vision and Pattern Recognition}, pages 18039--18048, 2024{\natexlab{a}}.

\bibitem[Matsuki et~al.(2024{\natexlab{b}})Matsuki, Murai, Kelly, and Davison]{Matsuki:Murai:etal:CVPR2024}
Hidenobu Matsuki, Riku Murai, Paul H.~J. Kelly, and Andrew~J. Davison.
\newblock {G}aussian {S}platting {SLAM}.
\newblock In \emph{Proceedings of the IEEE/CVF Conference on Computer Vision and Pattern Recognition}, 2024{\natexlab{b}}.

\bibitem[Mildenhall et~al.(2021)Mildenhall, Srinivasan, Tancik, Barron, Ramamoorthi, and Ng]{mildenhall2021nerf}
Ben Mildenhall, Pratul~P Srinivasan, Matthew Tancik, Jonathan~T Barron, Ravi Ramamoorthi, and Ren Ng.
\newblock Nerf: Representing scenes as neural radiance fields for view synthesis.
\newblock \emph{Communications of the ACM}, 65\penalty0 (1):\penalty0 99--106, 2021.

\bibitem[Mur-Artal and Tard{\'o}s(2017)]{mur2017orb}
Raul Mur-Artal and Juan~D Tard{\'o}s.
\newblock Orb-slam2: An open-source slam system for monocular, stereo, and rgb-d cameras.
\newblock \emph{IEEE transactions on robotics}, 33\penalty0 (5):\penalty0 1255--1262, 2017.

\bibitem[Mur-Artal et~al.(2015)Mur-Artal, Montiel, and Tardos]{mur2015orb}
Raul Mur-Artal, Jose Maria~Martinez Montiel, and Juan~D Tardos.
\newblock Orb-slam: a versatile and accurate monocular slam system.
\newblock \emph{IEEE transactions on robotics}, 31\penalty0 (5):\penalty0 1147--1163, 2015.

\bibitem[Palazzolo et~al.(2019)Palazzolo, Behley, Lottes, Gigu\`ere, and Stachniss]{palazzolo2019iros}
E. Palazzolo, J. Behley, P. Lottes, P. Gigu\`ere, and C. Stachniss.
\newblock {ReFusion: 3D Reconstruction in Dynamic Environments for RGB-D Cameras Exploiting Residuals}.
\newblock 2019.

\bibitem[Qin et~al.(2018)Qin, Li, and Shen]{qin2018vins}
Tong Qin, Peiliang Li, and Shaojie Shen.
\newblock Vins-mono: A robust and versatile monocular visual-inertial state estimator.
\newblock \emph{IEEE Transactions on Robotics}, 34\penalty0 (4):\penalty0 1004--1020, 2018.

\bibitem[Rosinol et~al.(2023)Rosinol, Leonard, and Carlone]{rosinol2023nerf}
Antoni Rosinol, John~J Leonard, and Luca Carlone.
\newblock Nerf-slam: Real-time dense monocular slam with neural radiance fields.
\newblock In \emph{2023 IEEE/RSJ International Conference on Intelligent Robots and Systems (IROS)}, pages 3437--3444. IEEE, 2023.

\bibitem[Rusinkiewicz and Levoy(2001)]{rusinkiewicz2001efficient}
Szymon Rusinkiewicz and Marc Levoy.
\newblock Efficient variants of the icp algorithm.
\newblock In \emph{Proceedings third international conference on 3-D digital imaging and modeling}, pages 145--152. IEEE, 2001.

\bibitem[Sabour et~al.(2023)Sabour, Vora, Duckworth, Krasin, Fleet, and Tagliasacchi]{sabour2023robustnerf}
Sara Sabour, Suhani Vora, Daniel Duckworth, Ivan Krasin, David~J Fleet, and Andrea Tagliasacchi.
\newblock Robustnerf: Ignoring distractors with robust losses.
\newblock In \emph{Proceedings of the IEEE/CVF Conference on Computer Vision and Pattern Recognition}, pages 20626--20636, 2023.

\bibitem[Schops et~al.(2019)Schops, Sattler, and Pollefeys]{schops2019bad}
Thomas Schops, Torsten Sattler, and Marc Pollefeys.
\newblock Bad slam: Bundle adjusted direct rgb-d slam.
\newblock In \emph{Proceedings of the IEEE/CVF Conference on Computer Vision and Pattern Recognition}, pages 134--144, 2019.

\bibitem[Segal et~al.(2009)Segal, Haehnel, and Thrun]{segal2009generalized}
Aleksandr Segal, Dirk Haehnel, and Sebastian Thrun.
\newblock Generalized-icp.
\newblock In \emph{Robotics: science and systems}, page 435. Seattle, WA, 2009.

\bibitem[Sturm et~al.(2012)Sturm, Engelhard, Endres, Burgard, and Cremers]{sturm12iros}
J. Sturm, N. Engelhard, F. Endres, W. Burgard, and D. Cremers.
\newblock A benchmark for the evaluation of rgb-d slam systems.
\newblock In \emph{Proc. of the International Conference on Intelligent Robot Systems (IROS)}, 2012.

\bibitem[Sumner et~al.(2007)Sumner, Schmid, and Pauly]{10.1145/1276377.1276478}
Robert~W. Sumner, Johannes Schmid, and Mark Pauly.
\newblock Embedded deformation for shape manipulation.
\newblock \emph{ACM Trans. Graph.}, 26\penalty0 (3):\penalty0 80–es, 2007.

\bibitem[Sun et~al.(2024)Sun, Jiao, Li, Zhang, Zhao, and Xing]{sun20243dgstream}
Jiakai Sun, Han Jiao, Guangyuan Li, Zhanjie Zhang, Lei Zhao, and Wei Xing.
\newblock 3dgstream: On-the-fly training of 3d gaussians for efficient streaming of photo-realistic free-viewpoint videos.
\newblock In \emph{Proceedings of the IEEE/CVF Conference on Computer Vision and Pattern Recognition (CVPR)}, pages 20675--20685, 2024.

\bibitem[Teed and Deng(2020{\natexlab{a}})]{teed2020raft}
Zachary Teed and Jia Deng.
\newblock Raft: Recurrent all-pairs field transforms for optical flow.
\newblock In \emph{Computer Vision--ECCV 2020: 16th European Conference, Glasgow, UK, August 23--28, 2020, Proceedings, Part II 16}, pages 402--419. Springer, 2020{\natexlab{a}}.

\bibitem[Teed and Deng(2020{\natexlab{b}})]{teed2020raftrecurrentallpairsfield}
Zachary Teed and Jia Deng.
\newblock Raft: Recurrent all-pairs field transforms for optical flow, 2020{\natexlab{b}}.

\bibitem[Wang and Liao(2024)]{wang2024yolov9}
Chien-Yao Wang and Hong-Yuan~Mark Liao.
\newblock {YOLOv9}: Learning what you want to learn using programmable gradient information.
\newblock 2024.

\bibitem[Wu et~al.(2024)Wu, Yi, Fang, Xie, Zhang, Wei, Liu, Tian, and Wang]{Wu_2024_CVPR}
Guanjun Wu, Taoran Yi, Jiemin Fang, Lingxi Xie, Xiaopeng Zhang, Wei Wei, Wenyu Liu, Qi Tian, and Xinggang Wang.
\newblock 4d gaussian splatting for real-time dynamic scene rendering.
\newblock In \emph{Proceedings of the IEEE/CVF Conference on Computer Vision and Pattern Recognition (CVPR)}, pages 20310--20320, 2024.

\bibitem[Xu et~al.(2025)Xu, Jiang, Xiao, Feng, and Zhang]{xu2025dg}
Yueming Xu, Haochen Jiang, Zhongyang Xiao, Jianfeng Feng, and Li Zhang.
\newblock Dg-slam: Robust dynamic gaussian splatting slam with hybrid pose optimization.
\newblock \emph{Advances in Neural Information Processing Systems}, 37:\penalty0 51577--51596, 2025.

\bibitem[Yu et~al.(2018)Yu, Liu, Liu, Xie, Yang, Wei, and Fei]{yu2018ds}
Chao Yu, Zuxin Liu, Xin-Jun Liu, Fugui Xie, Yi Yang, Qi Wei, and Qiao Fei.
\newblock Ds-slam: A semantic visual slam towards dynamic environments.
\newblock In \emph{2018 IEEE/RSJ international conference on intelligent robots and systems (IROS)}, pages 1168--1174. IEEE, 2018.

\bibitem[Yugay et~al.(2023)Yugay, Li, Gevers, and Oswald]{yugay2023gaussianslam}
Vladimir Yugay, Yue Li, Theo Gevers, and Martin~R. Oswald.
\newblock Gaussian-slam: Photo-realistic dense slam with gaussian splatting, 2023.

\bibitem[Zhang et~al.(2019)Zhang, Li, Bi, Zheng, Wang, Huang, Luo, Xu, and Gao]{zhang2019ppgnet}
Ziheng Zhang, Zhengxin Li, Ning Bi, Jia Zheng, Jinlei Wang, Kun Huang, Weixin Luo, Yanyu Xu, and Shenghua Gao.
\newblock Ppgnet: Learning point-pair graph for line segment detection.
\newblock In \emph{Proceedings of the IEEE/CVF Conference on Computer Vision and Pattern Recognition}, pages 7105--7114, 2019.

\bibitem[Zheng et~al.(2013)Zheng, Kuang, Sugimoto, Astrom, and Okutomi]{zheng2013revisiting}
Yinqiang Zheng, Yubin Kuang, Shigeki Sugimoto, Kalle Astrom, and Masatoshi Okutomi.
\newblock Revisiting the pnp problem: A fast, general and optimal solution.
\newblock In \emph{Proceedings of the IEEE International Conference on Computer Vision}, pages 2344--2351, 2013.

\bibitem[Zhu et~al.(2024)Zhu, Fang, Li, Yan, Xu, Yuen, and Li]{zhu2024robust}
Zunjie Zhu, Youxu Fang, Xin Li, Chengang Yan, Feng Xu, Chau Yuen, and Yanyan Li.
\newblock Robust gaussian splatting slam by leveraging loop closure.
\newblock \emph{arXiv preprint arXiv:2409.20111}, 2024.

\bibitem[Zwicker et~al.(2002)Zwicker, Pfister, Van~Baar, and Gross]{zwicker2002ewa}
Matthias Zwicker, Hanspeter Pfister, Jeroen Van~Baar, and Markus Gross.
\newblock Ewa splatting.
\newblock \emph{IEEE Transactions on Visualization and Computer Graphics}, 8\penalty0 (3):\penalty0 223--238, 2002.

\end{thebibliography}
